\documentclass[runningheads]{llncs}

\usepackage{eccv}

\usepackage{eccvabbrv}

\usepackage{graphicx}
\usepackage{booktabs}

\usepackage{pifont}
\usepackage{titletoc}
\usepackage{algorithm}

\usepackage{algpseudocode}
\usepackage{wrapfig}

\usepackage{xcolor}
\usepackage{array} 
\usepackage{colortbl}

\usepackage{multirow}
\usepackage{subcaption}

\definecolor{mygray}{gray}{.92}

\newcolumntype{x}[1]{>{\centering\arraybackslash}p{#1pt}}
\newcolumntype{y}[1]{>{\raggedright\arraybackslash}p{#1pt}}
\newcolumntype{z}[1]{>{\raggedleft\arraybackslash}p{#1pt}}

\usepackage[accsupp]{axessibility}  

\usepackage{hyperref}

\usepackage{orcidlink}
\usepackage{bibunits}

\newcommand{\graycomment}[1]{\textcolor{gray}{\small\(\#\) #1}}

\usepackage[normalem]{ulem}

\begin{document}


\title{FedOT: Ownership Verification and Leakage Tracing via Watermarks for Federated LDMs} 

\titlerunning{FedOT}

\author{Wenlong Cheng\inst{1}\orcidlink{0009-0005-2891-9523} \and
Yuan Gan\inst{2}$^{*}$\orcidlink{0000-0002-1380-787X} \and
Yunqiu Xu\inst{3}\orcidlink{0000-0002-2940-4805} \and
Jiaxu Miao\inst{4}$^{*}$\orcidlink{0000-0002-4238-8475}}

\authorrunning{W. Cheng et al.}

\institute{\textsuperscript{1}Northwest Normal University \ \ 
\textsuperscript{2}The University of Tokyo\\
\textsuperscript{3}National University of Singapore \ \ 
\textsuperscript{4}Sun Yat-sen University
\\
\email{2023222209@nwnu.edu.cn, \ y-gan@mi.t.u-tokyo.ac.jp,\\ imyunqiuxu@gmail.com, \ miaojx@hit.edu.cn}
}

\maketitle

\begingroup
\renewcommand{\thefootnote}{}
\footnotetext{* Corresponding authors.}
\endgroup

\begin{abstract}
Training Latent Diffusion Models (LDMs) within Federated Learning (FL) has attracted increasing attention due to its ability to combine the powerful generative capacity of LDMs with the privacy-preserving properties of FL.
However, FL requires sharing the global model with multiple participants, which risks unauthorized model distribution or resale by malicious clients.
While an intuitive approach is to adopt existing VAE-based watermarking techniques for LDMs in FL, this strategy falls short in addressing such threats due to two fundamental challenges:
(1) Existing methods support ownership verification but lack the ability to trace model leakage to a specific malicious client;
(2) VAE-based watermarks are vulnerable, as they can be removed simply by replacing the decoder with a clean counterpart.
In this paper, we propose FedOT, the first framework for ownership verification and leakage tracing in federated LDMs.
Specifically, to address the first challenge, we design a chunked watermark, where the first part is for ownership verification, and the second part is used for client identification.
Furthermore, to overcome the second challenge and secure the model against VAE replacement attack, we introduce Latent Vector Transformation (LVT), which strengthens the connection between the VAE and U-Net latent spaces by modifying the original latent distribution of the VAE.
Consequently, any attempt to replace the VAE for watermark removal leads to significant image quality degradation, making the LDM model unusable.
Extensive experiments demonstrate that FedOT achieves superior performance in both ownership verification and traceability.
Project page: \url{https://spyzixuan.github.io/FedOT/}.
\end{abstract}    
\section{Introduction}
\label{sec:intro}

\begin{figure}[t]
    \centering
    \includegraphics[width=\columnwidth]{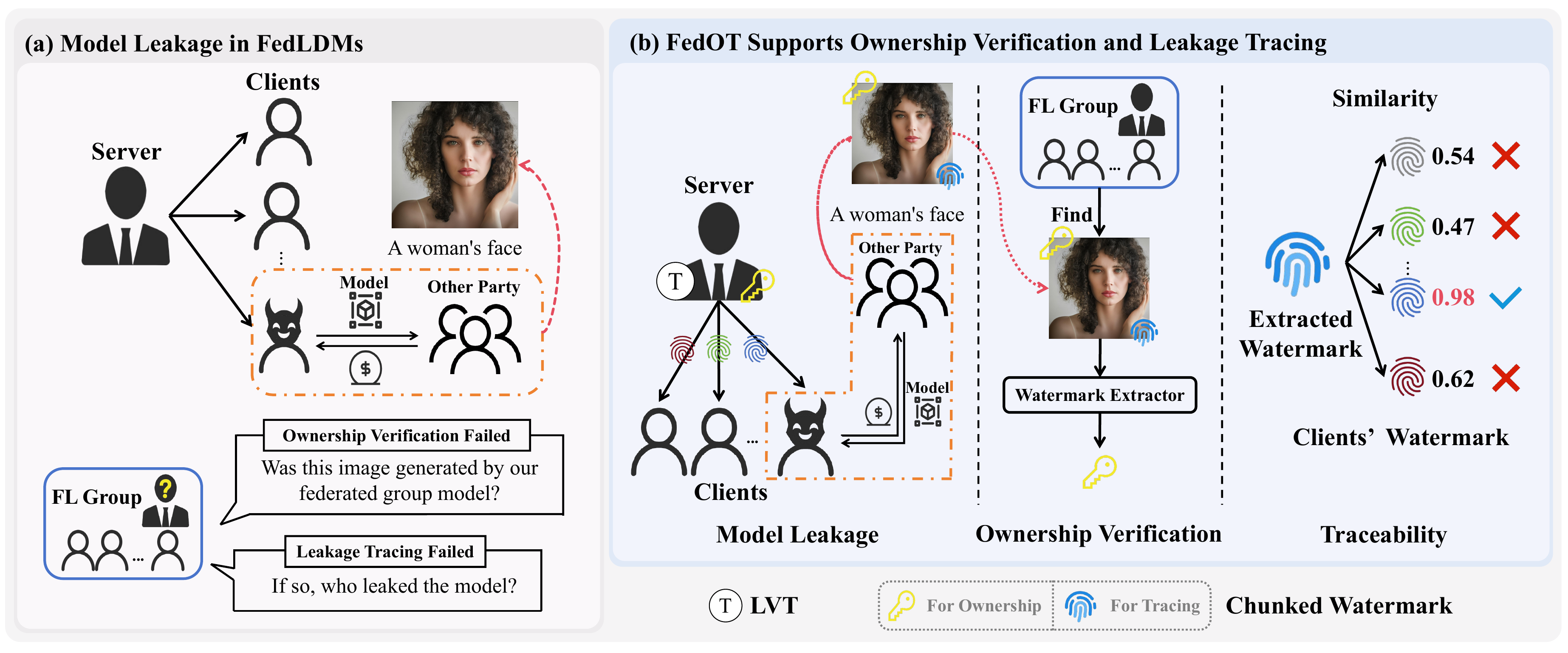}
\vspace{-2.25em}
    \caption{\textbf{Motivation of FedOT.} (a) Malicious clients leak the federated LDMs, verifying ownership or tracing the source of generated images becomes infeasible. (b) \textbf{FedOT supports ownership verification and malicious client tracing even if the leaked model is misused}, by using chunked watermark and Latent Vector Transformation (\textbf{LVT}).}
    \label{fig:application_scenario}
\vspace{-1em}
\end{figure}

Latent Diffusion Models (LDMs)~\cite{stable_diffusion, sdxl, balaji2022ediff, gu2022vector, nichol2021glide} have demonstrated remarkable capabilities in high-fidelity image synthesis and reshaped the AIGC landscape~\cite{cao2025survey,yang2021multiple}. Traditionally, these models rely on centralized training over massive open-source datasets.
To achieve privacy-preserving model training, federated Latent Diffusion Models (FedLDMs)~\cite{li2024feddiff, stanley2024phoenix, morafah2024stable, datastealing, liu2024iterative} synergize the strengths of LDMs with the privacy guarantees of Federated Learning (FL)~\cite{federated_learning}, enabling diffusion models to learn from distributed clients while preserving data privacy.

Despite successfully safeguarding data privacy, FedLDMs introduce a critical vulnerability concerning model ownership.
As illustrated in Fig.~\ref{fig:application_scenario}(a), the sharing of the global model among participants inevitably exposes the intellectual property to malicious clients.
They may distribute or resell the fine-tuned models without authorization.
Such leakage of models not only constitutes intellectual property infringement but also raises ethical concerns, including harmful content generation and privacy leakage~\cite{deepfake,deepfake1}.

While recent works~\cite{fedtracker, waffle} have explored copyright protection and traceability in FL, these solutions are exclusively designed for classification tasks.
The substantial architectural differences render them fundamentally inapplicable to complex generative models like LDMs.
To address this gap, we must look toward watermarking techniques specifically designed for LDMs~\cite{diffusetrace, ipwatermark, Stable_signature, tree-ring, yang2024gaussian, hu2025videoshield, wang2025sleepermark}.
However, adapting these techniques to FL requires careful consideration of the training process.
During LDM fine-tuning, clients typically only update the U-Net parameters, leaving the VAE entirely frozen.
Given this training characteristic, the VAE-based watermarking approach~\cite{Stable_signature} emerges as a highly intuitive and promising candidate for FL.
Despite its theoretical suitability, directly deploying this technique in a federated setting exposes two fundamental challenges:
\textbf{\ding{182} It can only verify that a model originates from the FL group (\ie, ownership verification), but cannot identify the specific malicious client responsible for leaking the model (\ie, leakage tracing)};
\textbf{\ding{183} The watermark can be easily removed without degrading the model's utility by replacing the watermark VAE decoder with a clean counterpart}.

To address these limitations, this paper proposes \textbf{FedOT, the first framework capable of both ownership verification and leakage tracing for FedLDMs}.
Rather than focusing on watermarking algorithms, our core design introduces a chunked watermark mechanism. Specifically, before distributing the global model, the server embeds different binary watermarks into the VAE decoder of each client’s LDM.
The first $r$ bits identify whether the model originates from the federated group. 
The remaining $n-r$ bits are unique to each client and are used to trace the source of the leak.
If a malicious client leaks the model, we can extract the watermark from the images generated by the leaked model.
An advantage of this design is that it runs the full tracing process only when the first $r$ bits confirm the model’s origin, making detection more efficient.

However, simply embedding the watermark into the VAE is vulnerable.
Malicious clients can effortlessly execute a zero-cost replacement attack by swapping the watermarked VAE with an available clean counterpart.
This effectively removes the watermark without degrading the quality of the generated images. 
To prevent such replacement attacks, we propose Latent Vector Transformation (LVT), a technique designed to tightly bind the VAE and U-Net components. Rather than modifying the watermarking algorithm, LVT proactively alters the latent space distribution of the VAE. Consequently, the U-Net is forced to gradually adapt to this modified distribution during the federated training process.

To balance the binding strength and generation quality, we explore three types of LVT strategies, namely translation, mirror, and negative, which modify the latent distribution to secure our framework against replacement attacks.

We systematically analyze the distribution shift of the three LVT strategies. Our experiments demonstrate that the negative transformation is the optimal choice considering the performance and component binding strength.
Overall, our main contributions can be summarized as follows:
\begin{itemize}
\item  
To the best of our knowledge, we 
propose the first framework, FedOT, that enables both ownership verification and leakage tracing for FedLDMs, filling a critical gap in existing research.

\item 
We introduce a VAE latent-space transformation, ensuring that the VAE remains compatible only with the U-Net trained in FL. Our proposed LVT establishes a strong dependency between model components: any attempt to remove the watermark by replacing the VAE incurs severe degradation in image quality, thereby deterring malicious clients from doing so.

\item 
We conduct experiments under various common watermark removal attacks, such as VAE replacement attack, model purification attack, and image attacks. The results show that FedOT provides a reliable mechanism for ownership verification and leakage tracing for FedLDMs.

\end{itemize}
\section{Related Work}
\label{sec:related}

\noindent\textbf{Federated Learning (FL).} FL enables multiple clients to collaboratively train models without sharing raw data, thereby preserving privacy. In a typical client-server architecture~\cite{yangfederated,safelearn}, each client updates the model with local data, while the server aggregates these updates using FedAvg~\cite{federated_learning} to form a global model. We adopt FL to train Latent Diffusion Models~\cite{stable_diffusion} for decentralized generative modeling, where only the U-Net parameters are uploaded and aggregated each round, reducing communication costs. However, FL also introduces new challenges for copyright protection, as the global model is accessible to all participating clients.


\vspace{0.3em}\noindent\textbf{Diffusion Models.}
Denoising Diffusion Probabilistic Models (DDPMs)~\cite{ddpm} generate data through iterative noise addition and removal. Latent Diffusion Models (LDMs)~\cite{stable_diffusion} improve efficiency by performing this process in a low-dimensional latent space, where images are encoded, denoised, and reconstructed by a decoder, or synthesized from a Gaussian prior via DDIM~\cite{ddim}. Diffusion models have achieved remarkable success in text-to-image generation~\cite{sd_clip_latent,video_sdm,gu2022vector,zhou2026bidedpo,balaji2022ediff} and editing tasks~\cite{controlnet,dreambooth,xu2024gg,jia2026gas,instructpix2pix}. Stable Diffusion~\cite{stable_diffusion,sdxl,sd3scaling}, a widely adopted open-source implementation, has further driven research and applications. Despite their strong generative capabilities, LDMs typically rely on centralized training with large-scale data, raising significant privacy concerns~\cite{datastealing,liu2024iterative,gan2025silence,tian2025brainguard}.


\vspace{0.3em}\noindent\textbf{Digital Watermark.} Watermarking is crucial for model traceability and intellectual property protection, spanning data watermarking~\cite{wm_train_datas}, image watermarking~\cite{hidden,steganography,liu2025watermarking_one_for_all}, and model watermarking~\cite{Stable_signature,tree-ring,safe-sd,yang2024gaussian,ipwatermark,hu2025videoshield,wang2025sleepermark,xia2026echoes,ci2406wmadapter}. For LDM-based generative models, watermarks can be embedded in generated images, model parameters, or latent space, each with distinct tradeoffs in robustness and flexibility. In FL, server-side methods are preferred over client-side approaches~\cite{yang2023watermarking,liu2021secure,fedipr} to prevent malicious backdoor injection. WAFFLE~\cite{waffle} pioneered server-side FL watermarking for ownership verification, while FedTracker~\cite{fedtracker} further enables per-client traceability but is limited to classification models. We propose FedOT to fill the gap in watermark tracing and ownership verification for generative models in FL.

\section{Methodology}
\label{sec:method}

\subsection{FedOT Framework}
\label{sec:fedOT_framework}

\begin{figure*}[t]
    \centering
    \includegraphics[width=\textwidth]{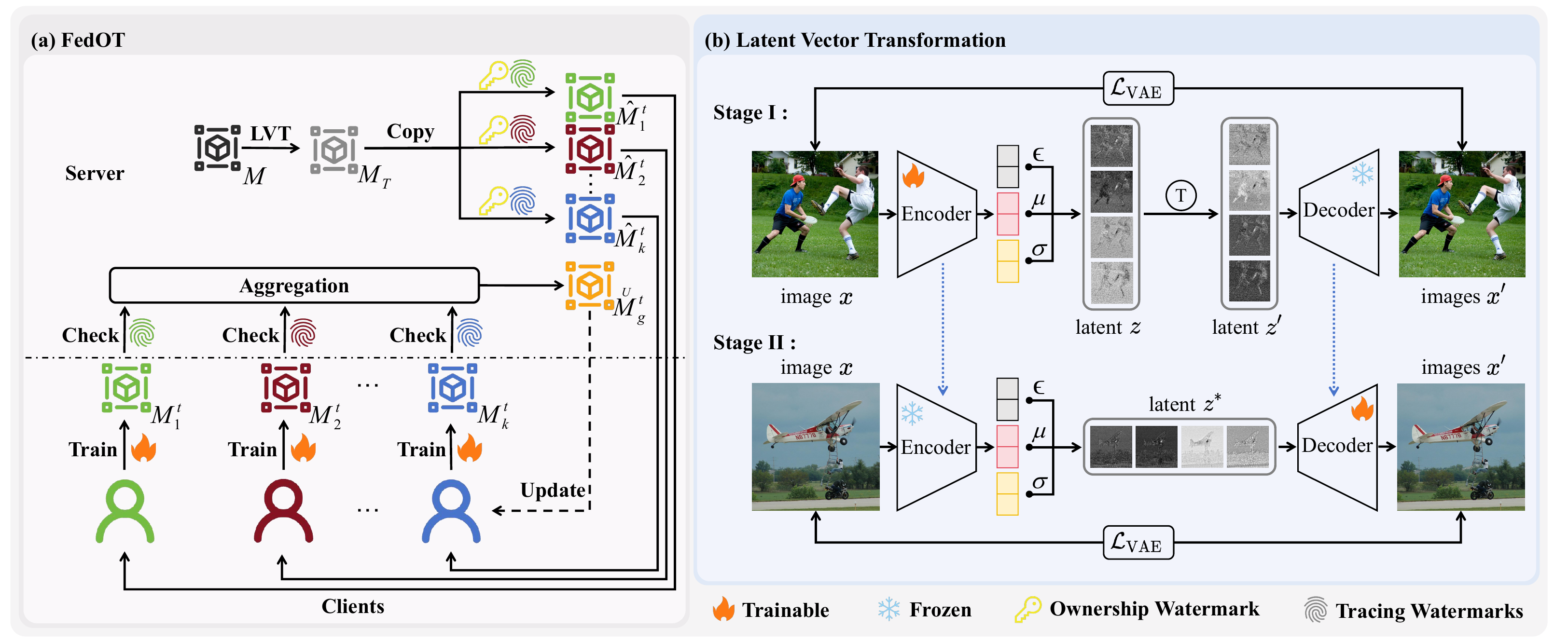}
\vspace{-2.25em}
    \caption{\textbf{Overview of FedOT.} 
(a) \textbf{FedOT Workflow.} The server applies LVT to the VAE, then embeds watermarks, and distributes model replicas to clients. Clients train the SD model locally and upload updates. The server verifies watermarks and aggregates only U-Net parameters.  
(b) \textbf{Details of LVT within FedOT.} Stage I: Transform latent vector $z$ to $z'$ and train the encoder. Stage II: Use the trained encoder to generate $z^*$ and adapt the decoder accordingly.
}
\vspace{-1.5em}
    \label{fig:pipeline}
\end{figure*}

As shown in Fig.~\ref{fig:pipeline}(a), our FedOT framework consists of two main components: Latent Vector Transformation (LVT) and watermark embedding.
FedOT is designed to collaboratively fine-tune Stable Diffusion\cite{stable_diffusion} on private distributed data in a federated setting.
To enhance intellectual property protection, the trusted server in FedOT first performs LVT training on the VAE module of the initial global model $M$, enabling the VAE to learn a transformed latent distribution and to produce a globally consistent latent representation. Based on this adapted global model $M_T$, the server then creates $K$ model replicas $\{M_i\}_{i=1}^K$, each injected with a unique $n$-bit watermark, resulting in the watermarked models $\{\hat{M}_i\}_{i=1}^K$. 
These watermarked models are then distributed to clients for local fine-tuning on their private data.
After each training round, clients upload their updated U-Net parameters and a subset of generated samples, which are aggregated on the server and redistributed for the next round of training.
A watermark extractor from the VAE-based method~\cite{Stable_signature} retrieves watermarks from the generated images. The overall FedOT procedure is illustrated in Algorithm~\ref{alg:FedOT_overall}.

If a malicious client leaks or resells the local LDM, the embedded watermark in generated images can be extracted to verify ownership and trace the source. The FL group first confirms ownership, then identifies the leaking client. More background details are provided in Appendix~\ref{sec:aa}.

\subsection{Watermark Design and Training} 
\label{sec:watermark-design}
\noindent\textbf{Watermark Design.} We propose a chunked watermark that supports both ownership verification and tracing. Specifically, the $n$-bit watermark is divided into two parts: the first $r$ bits for ownership verification, and the remaining $n-r$ bits for client tracing.
If the extracted watermark $\textbf{m}'$ matches the originally embedded watermark $\textbf{m}$ under the following condition, the image is verified as originating from the federated model:
\begin{equation}
\text{Verify}(\textbf{m}', \textbf{m}) =
\begin{cases}
\text{True}, & \text{Match}(\textbf{m}'_{1:r}, \textbf{m}_{1:r}) \geq \tau \\
\text{False}, & \text{otherwise,}
\end{cases}
\label{eq:verify}
\end{equation}
where $\text{Match}(\cdot)$ denotes the bit accuracy between two binary watermarks (Appendix~\ref{sub_sec:Bit_Accuracy_and_Detection.}).
Upon successful verification, the server proceeds to trace the specific client responsible for the leak:
\begin{equation}
j = \text{argmax}~ \text{Trace}(\textbf{m}'_{r+1:n}, \textbf{m}_{i,r+1:n}),
\label{eq:trace}
\end{equation}
where $j$ is the client index, indicating the $j$-th client leaked the model. This chunked watermark first verifies ownership via the shared $r$-bit prefix, and only performs full client identification using the $n-r$ suffix when necessary, reducing overhead in large-scale deployments.

\begin{algorithm*}[t]
\caption{Training Pipeline of FedOT}
\label{alg:FedOT_overall}
\begin{algorithmic}[1]
\item[\textbf{Input:}] 
Global model $M$, number of clients $K$, dataset $\{D_i\}_{i=1}^K$, watermark length $n$, verification length $r$, 
transformation type $T$, training steps $S$
\item[\textbf{Output:}] 
Final aggregated global model $M_g^T$
\Statex \graycomment{Train global model's VAE to learn the LVT transformation}
\State $E_T, D_T \gets LVTTraining(M,T)$ \Comment{Appendix~\ref{sec:CC}, Algorithm~\ref{alg:LVT_training}}
\vspace{0.3em}

\Statex \graycomment{Update the global model's VAE}
\State $M_T \gets UpdateGlobalModel(M,E_T,D_T)$ 
\vspace{0.3em}
\Statex \graycomment{Generate replicas with unique watermarks after LVT}
\State $\{\hat{M}_i\}_{i=1}^K \gets WatermarkEmbedding(M_T,K,n,r)$
\Comment{Appendix~\ref{sec:Watermark_Design_and_Training_Details}, Algorithm~\ref{alg:embed_watermark}}

\vspace{0.3em}

\Statex \graycomment{Perform federated training using watermarked model replicas}
\State $M_g^T \gets FederatedTrainingSD(\{\hat{M}_i\}_{i=1}^K,\{D_i\}_{i=1}^K,S)$
\Comment{Appendix~\ref{sec:aa}, Algorithm~\ref{alg:federated_sd}}

\vspace{0.3em}
\State \Return $M_g^T$
\end{algorithmic}
\end{algorithm*}

\vspace{0.3em}\noindent\textbf{Watermark Training.} During LDMs training, only the U-Net is updated while the VAE is frozen~\cite{dreambooth, han2023svdiff}. To ensure the embedded watermark remains unaffected during FL, inspired by Stable Signature~\cite{Stable_signature}, we embed the watermark into the decoder of the VAE. Before FL training, the trusted server embeds a watermark into the VAE decoder for each client. This watermark training is performed only once per client and does not need to be repeated in each communication round. To embed a unique watermark for each client, the server adopts the watermark extractor from Stable Signature to guide the training of the VAE decoder.
Specifically, we use a publicly available dataset~\cite{coco} to generate latent vectors $z$ via a VAE encoder.
The decoder then reconstructs the image $\hat{x}$ from $z$, and $\hat{x}$ is input to the watermark extractor to predict the embedded watermark $\textbf{m}'$. 
The Binary Cross-Entropy (BCE) loss is applied between $\textbf{m}'$ and the target watermark $\textbf{m}$ to ensure successful embedding:
\begin{equation}
\mathcal{L}_{m} = - \sum_{i=1}^{k} \left[ \textbf{m}_i \cdot \log \sigma(\textbf{m}'_i) + (1 - \textbf{m}_i) \cdot \log \left(1 - \sigma(\textbf{m}'_i)\right) \right].
\end{equation}

To preserve image quality, we additionally apply the Watson-VGG loss to encourage $\hat{x}$ to remain perceptually similar to the original input image $x$:
\begin{equation} 
\mathcal{L}_{i} = \text{Watson-VGG}\left(\hat{x}, x\right),
\end{equation}
where a weighted coefficient $\lambda_i$ is used to balance image fidelity and watermark bit accuracy, and the final objective can be formulated as:
\begin{equation} 
\mathcal{L}_{w} = \mathcal{L}_{m} + \lambda_i \cdot \mathcal{L}_{i}. 
\end{equation}

Fine-tuning the VAE decoder with this loss keeps the watermark intact despite U-Net updates, avoiding training interference throughout the entire optimization process. More training details are provided in Appendix~\ref{sec:Watermark_Design_and_Training_Details}.

\subsection{Latent Vector Transformation}
\label{sec:latent} 
Since the VAE is typically frozen during LDM fine-tuning, malicious clients can easily replace the watermarked VAE with an open-source version.
This effectively erases the embedded watermarks without incurring any degradation in image generation quality.
To prevent this, we propose Latent Vector Transformation (LVT) to establish a strong dependency between the VAE and U-Net in the latent space. 
Our key insight relies on the training mechanism of LDMs: the U-Net gradually adapts to the latent distribution produced by the VAE encoder.
Consequently, modifying this distribution forces the U-Net to shift toward a new latent space, subsequently causing it to "forget" the original pre-trained distribution. 
Exploiting this property, we fine-tune the global VAE to transform its latent space prior to federated training. This process effectively binds the subsequently trained U-Net to our modified VAE, rendering any decoder replacement attacks destructive to the model's utility.

\vspace{0.3em}\noindent\textbf{LVT Training.} Before watermark embedding, the server fine-tunes the VAE using a public dataset~\cite{coco} to modify the latent distribution in a controlled manner. We introduce a transformation $T$ to adjust latent vectors, ensuring the U-Net operates only on the transformed space.
As shown in Fig.~\ref{fig:pipeline}(b), the LVT process includes two stages. 
In stage I, the encoder is trained to learn the transformation $T$, while the decoder remains fixed. 
Given an image $x$, the VAE encoder maps it to a latent vector $z = E(x)$, which is then transformed into $z' = T(z)$. The decoder reconstructs the final image as $x' = D(z')$. To faithfully reconstruct the original image from $z'$, the encoder implicitly learns to approximate the transformation $T$ during training.
In stage II, the encoder is kept frozen, and the decoder is fine-tuned. Since the encoder has already learned the transformation $T$ during stage I, its current output $z^*$ differs from the original latent vector $z$. The decoder is trained to reconstruct images from the transformed latent space $z^*$ by implicitly learning the inverse transformation $T^{-1}$, allowing the decoder to adapt to the new latent space.

However, changes in the latent space can pose challenges for FL, as the U-Net needs to adapt to the new latent distribution. 
Therefore, it is critical to design LVT strategies with an optimal balance: they must induce a sufficient structural shift to break compatibility with clean VAEs, while minimizing any adverse impact on the final generative fidelity of the U-Net.

\vspace{0.3em}\noindent\textbf{Random Transformation.} The latent vector $z$ is typically sampled via the reparameterization trick from the encoder output mean $\mu$ and variance $\sigma$, formulated as $z = \mu + \sigma \cdot \epsilon$, where $\epsilon \sim \mathcal{N}(0, I)$.  Our objective is to fine-tune the VAE so that the latent space adopts a new structural transformation while preserving its Gaussian properties.

First, based on the properties of the Gaussian distribution, we consider adding a random normal distribution $\hat\epsilon$ to the latent vector $z$, yielding $z'=z+\hat\epsilon$. Under this transformation, $z'$ still follows a Gaussian distribution:
\begin{equation}
    z' \sim \mathcal{N}(\mu , I(\sigma^2+1)).
\end{equation}

However, results reveal that applying random Gaussian transformations significantly degrades the image reconstruction quality. 
We denote this method as \textbf{$\text{FedOT}_\text{rand}$} and include it as a baseline for comparison.
Our analysis indicates that the VAE has difficulty adapting random Gaussian transformations. 
\begin{wrapfigure}{r}{0.48\columnwidth}
    \centering
    \includegraphics[width=0.48\columnwidth]{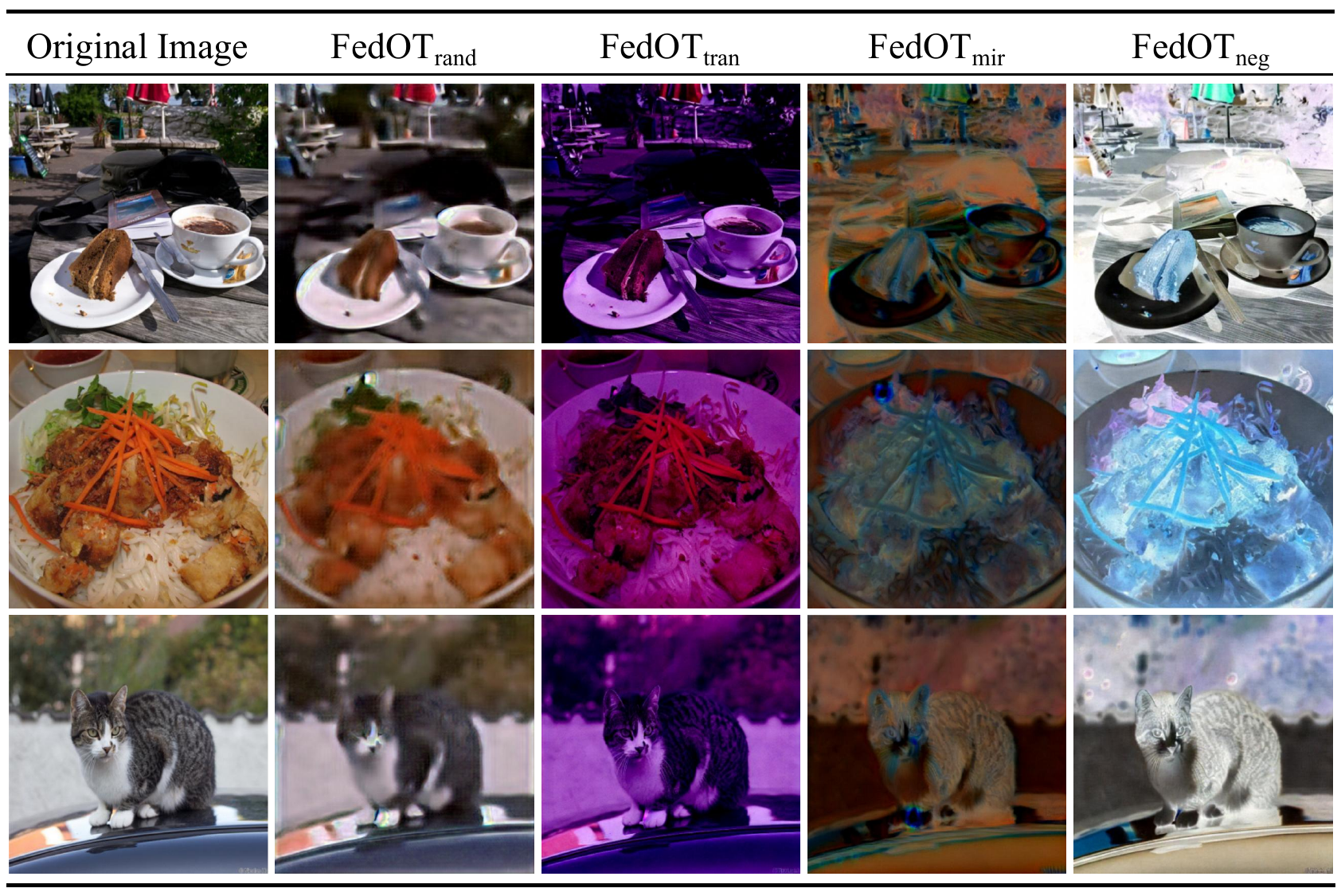}
    \vspace{-2em}
    \caption{VAE reconstruction results after training the encoder with different LVT transformations.}
    \label{fig:vae_latent}
    \vspace{-2.5em}
\end{wrapfigure}
Although standard Gaussian noise is added in each training iteration, the VAE cannot learn a consistent deterministic mapping to counteract this perturbation.
To seek a deterministic approach, we draw inspiration from recent findings~\cite{morita2025tkg_dm}, which preliminarily observed that applying constant shifts to the latent space can predictably alter generative attributes like color. We significantly advance this insight: rather than merely manipulating visual outputs, we formalize systematic, non-stochastic latent shifts as a mechanism to structurally bind LDM components against adversarial replacement attacks. Building on this novel perspective, we design and explore the following three stable LVT strategies.

\vspace{0.3em}\noindent\textbf{Translation Transformation.} Leveraging the linearity of Gaussian distributions, we apply a deterministic translation to the latent vector $z$, defined as $z' = z + c$. Under this operation, the transformed latent variable $z'$ follows:
\begin{equation}
    z' \sim \mathcal{N}(\mu + c, \sigma^2).
\end{equation}

We denote this method as \textbf{$\text{FedOT}_\text{tran}$}. This strategy shifts the distribution mean while preserving the original variance structure.
During Stage I fine-tuning, the VAE encoder learns the translation T, effectively mapping inputs to a globally shifted latent space. Subsequently, in Stage II, the frozen encoder produces shifted outputs, forcing the decoder to adapt its reconstruction process to align with this translated space.
As illustrated in Fig.~\ref{fig:vae_latent}, the $\text{FedOT}_\text{tran}$ method introduces severe color shifts after training the VAE encoder, which significantly degrades the quality of the generated images.

\vspace{0.3em}\noindent\textbf{Mirror Transformation.} For Gaussian distributions, applying a mirror operator $z' = -z$ to the latent variable $z$ mirrors the probability distribution across the origin. Under this transformation, the new latent variable $z'$ follows:
\begin{equation}
    z' \sim \mathcal{N}(-\mu, \sigma^2),
\end{equation}
which preserves the Gaussian variance structure while inverting the mean. We designate this approach as \textbf{$\text{FedOT}_\text{mir}$}.  
Compared with $\text{FedOT}_\text{tran}$, the mirror transformation performs a symmetric reflection of the entire latent space. During Stage I, the VAE encoder learns to map inputs to this inverted latent representation. Subsequently, in Stage II, the decoder attempts to adapt its reconstruction to align with this mirrored space.

We observe that mirroring the latent space does not correspond to a simple semantic inversion (\eg, color negation) in the pixel domain. Instead, it introduces severe distortions, particularly in color fidelity and fine-grained textures.
As shown in Fig.~\ref{fig:vae_latent}, this process often leads to color inversion and blurring, significantly degrading the image quality after training the VAE encoder.

\vspace{0.3em}\noindent\textbf{Negative Transformation.} To preserve high-frequency details while enforcing a structural shift, we study pixel-domain inversions for latent transformation.
Unlike direct latent operations, this strategy operates on the input data manifold, thereby implicitly reshaping the latent distribution while maintaining the Gaussian prior.
Specifically, we introduce a deterministic pixel-wise inversion, denoted as $x^{-} = 1 - x$, and compel the VAE to adapt this mapping for normalized inputs. During Stage I, the encoder is trained to map the standard input $x$ to a latent representation that corresponds to its negative counterpart $x^{-}$. Conversely, the decoder is tasked with reconstructing the original positive image $x$ from this inverted latent code.
This approach effectively implements a pixel negation within the latent space. We designate this method \textbf{$\text{FedOT}_\text{neg}$}. As shown in Fig.~\ref{fig:vae_latent}, compared to the reflection in {$\text{FedOT}_\text{mir}$}, this pixel-guided strategy preserves more structural details and avoids the edge blurring artifacts.

Crucially, these latent transformations render the watermarked VAE necessary to the generative pipeline. Any attempt by a malicious client to replace the VAE with a clean counterpart will inevitably fail, as the U-Net has been trained to depend on the modified latent distribution. Without knowledge of the specific transformation parameters, such unauthorized replacements result in a mismatch between the latent space and the U-Net, leading to severe degradation of image synthesis quality. More implementation and training details are provided in Appendix~\ref{sec:CC}.

\section{Experiments}

\subsection{Experimental Setup}

\noindent\textbf{Datasets.} We leverage the COCO2017\cite{coco} dataset to train the latent vector transformations of the VAE and embed the watermark, while utilizing LAION-10K\cite{laion_10k} to simulate private data for federated fine-tuning of Stable Diffusion. Specifically, we employ a subset of COCO2017 comprising 10,000 diverse images (resized to 512×512) spanning 80 categories. For the federated scenario, we curate LAION-10K, a subset of LAION~\cite{laion} containing 10,000 high-quality image-text pairs. Consistent with the protocol in IET~\cite{liu2024iterative}, these images are resized to 256×256 for efficient local training.

\vspace{0.3em}\noindent\textbf{Metrics.} To comprehensively evaluate the performance of our proposed FedOT framework, we employ multiple widely used metrics. We use FID~\cite{fid}, SSIM~\cite{ssim}, and PSNR to evaluate VAE reconstruction. FID and CLIP-Score~\cite{clip} assess Stable Diffusion generation quality. 
Detection Rate and Bit Accuracy are used to reflect the reliability of watermark extraction and its effectiveness.

\vspace{0.3em}\noindent\textbf{Implementation Details.} Training diffusion models in a federated setting is challenging, as the U-Net parameters are continuously updated and latent-space-based watermarks are unsuited for this task. Therefore, we adopt a VAE-based Stable Signature~\cite{Stable_signature} as our baseline. For training the LVT on the VAE, we set \(\lambda_{KL} = 10^{-8}\) and the translation coefficient for $\text{FedOT}_\text{tran}$ to 11. During watermark training, $\lambda_i = 0.2$. For federated fine-tuning, we deploy Stable Diffusion v2.1~\cite{stable_diffusion} across $K=5$ clients using the LAION-10K dataset, partitioned in an independent and identically distributed (i.i.d.) manner. Each client model undergoes local fine-tuning for 15 epochs, with 2,000 steps per epoch. The watermark length is set to $n=48$ bits, partitioned into a prefix of $r=16$ bits for global ownership verification and a suffix of 32 bits for precise client tracking. The detection threshold $\tau$ is set to 0.69, resulting in a FPR of 0.1\%. We define attack failure as post-attack FID exceeding the original SD baseline (i.e., 22.99 FID as in Table 1), meaning all FL fine-tuning benefits are destroyed. With 4 RTX 4090, LVT training takes $\sim$24 h; training an FL client requires $\sim$7.5 h/GPU (15 epochs); training a watermarked VAE takes $\sim$5 min/GPU.

\subsection{Main Results}
 
\textbf{Quantitative Analysis of Federated Fine-Tuning.}
Table~\ref{tab:main_table} reports the results of fine-tuning Stable Diffusion under the FedOT framework.  
The Original SD baseline refers to the pre-trained Stable Diffusion V2 model evaluated on the LAION-10K validation set using FID and CLIP-Score.
We adapt Stable Signature~\cite{Stable_signature}, originally designed for centralized training, to the federated setting as a comparison baseline, denoted as Stable Signature*.
FedOT$_\text{w/o LVT}$ represents our method without applying the Latent Vector Transformation (LVT) strategy.

As shown in Table~\ref{tab:main_table}, the watermarks in both the baseline Stable Signature* and $\text{FedOT}_\text{w/o LVT}$ are not detected after the replacement attack, while the image quality remains almost unaffected. This indicates that the attack removes the watermark at little cost. 
In contrast, our method with the proposed LVT strategy experiences a noticeable degradation in image quality after the attack. 
This demonstrates LVT's effectiveness in preventing watermark removal attacks.

\begin{table*}[t]
\centering
\caption{Generation quality and comparison with Stable Signature* on 256×256 images and 48-bit watermarks.
The left table reports results before the VAE replacement attack, while the right table shows performance after the attack.
Changes in generated image quality before and after the attack are indicated as \textcolor{green}{(green)} for increases and \textcolor{red}{(red)} for decreases. Stable Signature* refers to applying Stable Signature watermarking within federated learning.}
\vspace{-1em}
\resizebox{\textwidth}{!}{
\begin{tabular}{c||cccc|cccc}
\hline
\rowcolor[HTML]{EFEFEF} 
\cellcolor[HTML]{EFEFEF} & \multicolumn{4}{c|}{\textbf{Original}} & \multicolumn{4}{c}{\textbf{VAE Replacement Attack}} \\ 
\cline{2-9} 
\rowcolor[HTML]{EFEFEF} 
\multirow{-2}{*}{\textbf{Method}} & FID$\downarrow$ & CLIP-Score$\uparrow$ & Detection$\uparrow$ & Bit Acc$\uparrow$ 
                         & FID$\uparrow$ & CLIP-Score$\downarrow$ & Detection$\uparrow$ & Bit Acc$\uparrow$ \\ \hline \hline

\multicolumn{1}{c||}{Original SD~\cite{stable_diffusion}} & 22.99 & 0.316 & -- & -- & -- & -- & -- & -- \\
 Stable Signature*~\cite{Stable_signature} & 16.412 & 0.314 & 0.994 & 0.965 & 16.282 \textcolor{red}{(-0.130)}  & 0.312 \textcolor{red}{(-0.002)} & 0.000 & 0.543 \\
 FedOT$_\text{w/o LVT}$ & 16.687 & 0.313 & 0.998 & 0.972 & 17.054 \textcolor{green}{(+0.367)} & 0.311 \textcolor{red}{(-0.002)} & 0.000 & 0.494 \\
 $\text{FedOT}_\text{rand}$ & 35.585 & 0.284 & 0.999 & 0.980 & 72.810 \textcolor{green}{(+37.225)} & 0.260 \textcolor{red}{(-0.024)} & 0.000 & 0.536 \\
 \hline
 $\text{FedOT}_\text{tran}$ & 22.427 & 0.301 & 0.988 & 0.950 & 92.497 \textcolor{green}{(+70.070)} & 0.268 \textcolor{red}{(-0.033)} & 0.000 & 0.475 \\
 $\text{FedOT}_\text{mir}$ & 21.475 & 0.293 & 0.962 & 0.926 & 70.622 \textcolor{green}{(+49.147)} & 0.229 \textcolor{red}{(-0.064)} & 0.000 & 0.522 \\
 $\text{FedOT}_\text{neg}$ & 20.367 & 0.295 & 0.954 & 0.922 & 40.537 \textcolor{green}{(+20.170)} & 0.272 \textcolor{red}{(-0.023)} & 0.000 & 0.524 \\ \hline
\end{tabular}
}
\vspace{-0.5em}
\label{tab:main_table}
\end{table*}

\begin{figure*}[t]
\centering
\includegraphics[width=\textwidth]{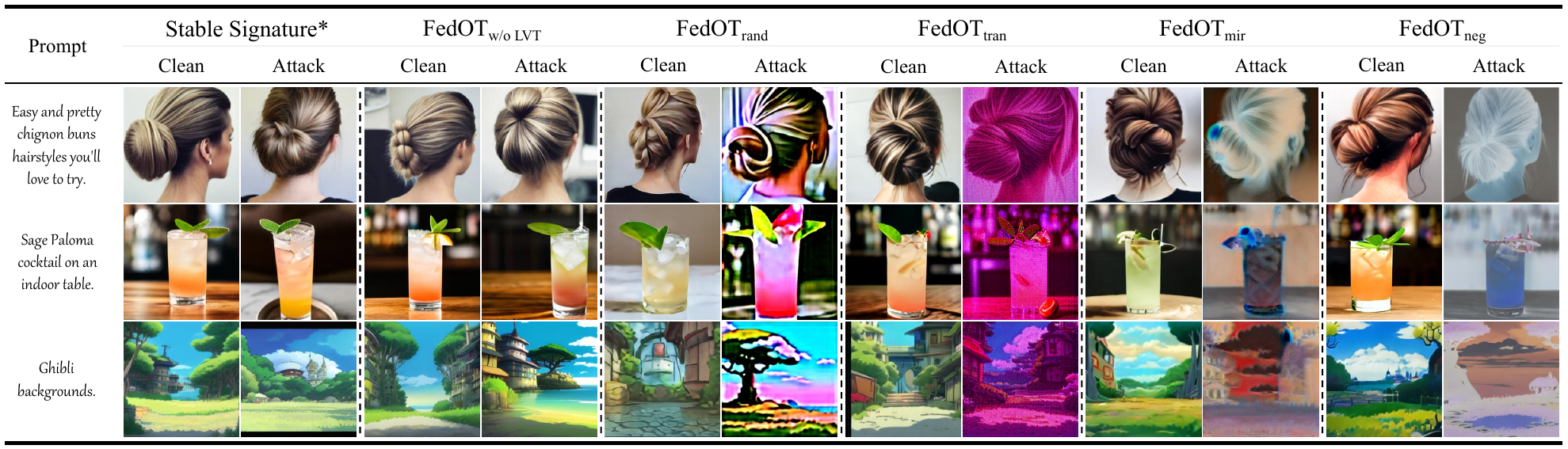}
\vspace{-2em}
    \caption{Comparison between different FedOT methods and Stable Signature*. \texttt{"}Clean\texttt{"} represents the results without VAE replacement attacks, while \texttt{"}Attack\texttt{"} represents the results after VAE replacement attacks. This figure corresponds to Table~\ref{tab:main_table}.}
\vspace{-1.5em}
\label{fig:results}
\end{figure*}

\vspace{0.3em}\noindent\textbf{Performance of LVT under FedOT.} We analyze different LVT strategies in Table~\ref{tab:main_table}.
$\text{FedOT}_\text{rand}$ exhibits strong resistance to VAE replacement (+37.225) but suffers from severe degradation in generation quality (FID: 35.585). 
This supports our view that the VAE latent space is difficult to adapt to a randomly sampled Gaussian distribution.
$\text{FedOT}_\text{tran}$ demonstrates strong resistance to VAE replacement (+70.070), but with lower image quality (FID: 22.427). This result reveals a critical trade-off: introducing a large translation coefficient induces a significant structural shift in the latent space, which effectively binds the components against replacement attacks but inevitably degrades generative fidelity.
In comparison, $\text{FedOT}_{mir}$ exhibits similarly robust binding capability (+49.147) and offers improved generation quality (FID: 21.475). However, a critical observation is its severe impact on semantic consistency: after the replacement attack, the CLIP-Score drops significantly (-0.064). This indicates that the mirror transformation disrupts the semantic alignment between the text prompts and the generated images, rendering the stolen model practically useless for content creation.
Finally, $\text{FedOT}_{neg}$ achieves the optimal balance, maintaining superior generation quality (FID: 20.367) while ensuring strong defense (+20.170). By leveraging pixel-domain inversion, this strategy preserves necessary high-frequency details while enforcing the latent shift. Unlike the translation and mirror transformations, it sustains a more consistent semantic structure, proving that carefully designed transformations can enhance watermark protection without sacrificing quality.

\vspace{0.3em}\noindent\textbf{Visual Impact of Attacks.} Fig.~\ref{fig:results} visually compares the generative outputs under VAE replacement attacks. Consistent with Table~\ref{tab:main_table}, Stable Signature* and $\text{FedOT}_\text{w/o~LVT}$ produce high-fidelity images post-attack, confirming their vulnerability to watermark removal without utility loss. In contrast, FedOT variants with LVT show visible degradation, demonstrating defense effectiveness. $\text{FedOT}_\text{tran}$ induces perceptible global color shifts, while 
$\text{FedOT}_\text{neg}$ manifests a clear luminance inversion. Notably, $\text{FedOT}_\text{mir}$ results in severe, unstructured distortions that disrupt semantic coherence, rendering the generated content visually unrecognizable and practically unusable for malicious actors.

\vspace{0.3em}\noindent\textbf{Ownership Verification and Tracing.}
As shown in Table~\ref{tab:Ownership verification and Tracing}, Stable Signature* achieves ownership verification but lacks client traceability.
In contrast, our FedOT framework supports both, due to its watermark design.
Among variants, $\text{FedOT}_\text{w/o LVT}$ performs best in both metrics.
However, this variant and Stable Signature* are vulnerable to replacement attacks, as shown in Table~\ref{tab:main_table}.
With the introduction of LVT, the performance of $\text{FedOT}_\text{tran}$, $\text{FedOT}_\text{mir}$, and $\text{FedOT}_\text{neg}$ slightly decreases but remains robust.
All variants maintain detection rates above 0.932 and bit accuracy over 0.91, demonstrating the effectiveness and necessity of our method.

\begin{table}[t]
\centering
\scriptsize
\renewcommand{\arraystretch}{1}
\caption{Comparison of Stable Signature* and the FedOT in the federated learning framework, focusing on ownership verification and tracing.}
\vspace{-1.25em}
\begin{tabular}{c||cc|cc}
\hline
\rowcolor[HTML]{EFEFEF} 
\cellcolor[HTML]{EFEFEF} & \multicolumn{2}{c|}{\textbf{Ownership}} & \multicolumn{2}{c}{\textbf{Tracing}} \\ 
\cline{2-5} 
\rowcolor[HTML]{EFEFEF} 
\multirow{-2}{*}{\textbf{Method}} & Detection$\uparrow$ & Bit Acc$\uparrow$ & Detection$\uparrow$ & Bit Acc$\uparrow$ \\ \hline \hline

\multicolumn{1}{c||}{Stable Signature*} & 0.994 & 0.965 & -- & -- \\ 

FedOT$_\text{w/o LVT}$ & 0.996 & 0.981 & 0.993 & 0.967 \\

$\text{FedOT}_\text{tran}$ & 0.983 & 0.962 & 0.975 & 0.945 \\

$\text{FedOT}_\text{mir}$ & 0.972 & 0.953 & 0.935 & 0.913 \\

$\text{FedOT}_\text{neg}$ & 0.960 & 0.947 & 0.932 & 0.910 \\ \hline
\end{tabular}
\vspace{-1em}
\label{tab:Ownership verification and Tracing}
\end{table}

\begin{table*}[t]
\centering
\caption{Performance of $\text{FedOT}_\text{tran}$ under varying translation coefficients $c$.}
\vspace{-1em}
\resizebox{\textwidth}{!}{
\begin{tabular}{c||cccc|cccc}
\hline
\rowcolor[HTML]{EFEFEF} 
\cellcolor[HTML]{EFEFEF} & \multicolumn{4}{c|}{\textbf{Original}} & \multicolumn{4}{c}{\textbf{VAE Replacement Attack}} \\ 
\cline{2-9} 
\rowcolor[HTML]{EFEFEF} 
\multirow{-2}{*}{\textbf{$\text{FedOT}_\text{tran}$}} & FID$\downarrow$ & CLIP-Score$\uparrow$ & Detection$\uparrow$ & Bit Acc$\uparrow$ 
                         & FID$\uparrow$ & CLIP-Score$\downarrow$ & Detection$\uparrow$ & Bit Acc$\uparrow$ \\ \hline \hline

\multicolumn{1}{c||}{$c=2$} & 22.656 & 0.303 & 0.959 & 0.940 & 21.232 \textcolor{red}{(-1.424)} & 0.305 \textcolor{green}{(+0.002)} & 0.000 & 0.507 \\
$c=5$ & 19.684 & 0.306 & 0.940 & 0.935 & 20.749 \textcolor{green}{(+1.065)} & 0.305 \textcolor{red}{(-0.001)} & 0.000 & 0.513 \\
$c=8$ & 21.461 & 0.305 & 0.959 & 0.939 & 23.752 \textcolor{green}{(+2.291)} & 0.299 \textcolor{red}{(-0.006)} & 0.000 & 0.512 \\
$c=11$ & 22.427 & 0.301 & 0.974 & 0.950 & 92.497 \textcolor{green}{(+70.070)} & 0.268 \textcolor{red}{(-0.033)} & 0.000 & 0.475 \\
 \hline
\end{tabular}
}
\vspace{-1.25em}
\label{tab:Translation Coefficient c}
\end{table*}

\begin{table*}[t]
\centering
\scriptsize
\setlength{\tabcolsep}{2pt}
\renewcommand{\arraystretch}{1}
    \caption{Trade-off between PSNR and Bit Accuracy under different $\lambda_i$}
\vspace{-1em}
\begin{tabular}{c||ccccc}
\hline
\rowcolor[HTML]{EFEFEF} 
\textbf{$\lambda_i$ for Fine-tuning} & 0.8 & 0.4 & 0.2 & 0.1 & 0.05 \\ \hline
PSNR $\uparrow$ & 34.793 & 34.223 & 33.611 & 32.953 & 29.842 \\
Bit Acc $\uparrow$ & 0.906 & 0.927 & 0.935 & 0.960 & 0.965 \\
\hline 
\end{tabular}
\vspace{-1em}
\label{tab:Coefficient lambda_i}
\end{table*}

\begin{table}[t]
\centering
\scriptsize
\setlength{\tabcolsep}{2pt}
\renewcommand{\arraystretch}{1}
    \caption{impact of different $r$ on ownership verification and tracing.}
\vspace{-1em}
\begin{tabular}{c||cc|cc}
\hline
\rowcolor[HTML]{EFEFEF} 
& \multicolumn{2}{c|}{\textbf{Ownership}} & \multicolumn{2}{c}{\textbf{Tracing}} \\ \cline{2-5}
\rowcolor[HTML]{EFEFEF}
\multirow{-2}{*}{\textbf{\textbf{$r$}-length}} 
& Detection $\uparrow$ & Bit Acc $\uparrow$ & Detection $\uparrow$ & Bit Acc $\uparrow$ \\ \cline{2-5} \hline \hline
$r = 8$ & 0.944 & 0.928 & 0.957 & 0.921 \\
$r = 16$ & {0.960} & 0.947 & {0.932} & {0.910} \\
$r = 24$ & 0.958 & {0.946} & 0.928 & 0.899 \\ \hline
\end{tabular}
\vspace{-0.5em}
  \label{tab:r_value_ablation}
\end{table}

\subsection{Ablation Studies}
\noindent\textbf{Translation Coefficient c.} We investigate the sensitivity of $\text{FedOT}_\text{tran}$ to the translation coefficient c. As expected, our results reveal a critical trade-off between generative fidelity and resistance to replacement attacks. Setting c to an excessively small or large value degrades the models' optimal performance. Interestingly, under a weak perturbation ($c=2$), replacing the watermarked VAE with a clean counterpart actually results in improvement in image quality. This suggests that too small latent shifts fail to enforce component binding. Conversely, as c increases, the binding becomes increasingly stringent. When $c=11$, the image quality degradation induced by a replacement attack surpasses even that of the $\text{FedOT}_\text{mir}$. These findings empirically demonstrate that a larger translation coefficient significantly amplifies the models' ability against attacks, but at the inevitable cost of generation quality.

\vspace{0.3em}\noindent\textbf{Impact of Weighting Coefficient $\lambda_i$.} We further perform an ablation study on the weighting coefficient $\lambda_i$. As shown in Table~\ref{tab:Coefficient lambda_i}, varying $\lambda_i$ introduces a direct tension between visual quality and watermark reliability. A larger $\lambda_i$ prioritizes the reconstruction loss, compelling the generated images to closely resemble the original inputs. However, this heavily penalizes the watermark embedding loss, resulting in significantly lower bit accuracies during message extraction. Conversely, decreasing $\lambda_i$ strengthens the watermark, ensuring bit accuracy.

\vspace{0.3em}\noindent\textbf{Different Lengths of the Chunked Watermark $r$.} In the FedOT framework, the watermark of length $n=48$ is split into two segments: the first $r$ bits are designated for ownership verification, while the remaining $n-r$ bits are assigned to client tracing. We explore how different values of $r$ affect both tasks by conducting ablation experiments, with results reported in Table~\ref{tab:r_value_ablation}. When $r = 8$, fewer bits are available for ownership verification, leading to weaker performance (Detection: 0.944, Bit Acc: 0.928). When $r = 24$, the tracing performance noticeably drops (Detection: 0.928, Bit Acc: 0.899) due to fewer bits being available for client encoding. The setting $r = 16$ achieves the best balance, with strong performance in both ownership and tracing. This configuration is therefore used as the default in all other experiments.

\subsection{Impact of Federated Conditions}
\noindent\textbf{FedOT with Different Numbers of Clients.}  
We evaluate the performance of FedOT with 5, 10, and 20 clients, as shown in Table~\ref{tab:different_clients_performance}. Overall, the generation quality remains relatively stable as the number of clients increases. Meanwhile, watermark robustness stays largely unaffected across all settings, further demonstrating the effectiveness of our LVT-based watermarking approach. In all cases, detection rates exceed 0.941, indicating that even in federated learning environments with diverse and heterogeneous client datasets, the number of clients has a limited impact on the ability to reliably detect and trace watermarks.

\vspace{0.3em}\noindent\textbf{FedOT under Non-i.i.d. Distributions.}  
To assess robustness under realistic conditions, we simulate non-i.i.d. settings using Dirichlet distributions with varying $\alpha$, as shown in Table~\ref{tab:non_iid_distribution}.  
With 5 clients, all FedOT variants maintain stable generation quality and watermark robustness across different $\alpha$ values.
In all cases, the detection remains above 0.957, indicating that data heterogeneity has little impact on watermark robustness.

\subsection{Purification Attack} 
\begin{wrapfigure}{r}{0.5\columnwidth}
    \centering
    \vspace{-2em}
    \includegraphics[width=\linewidth]{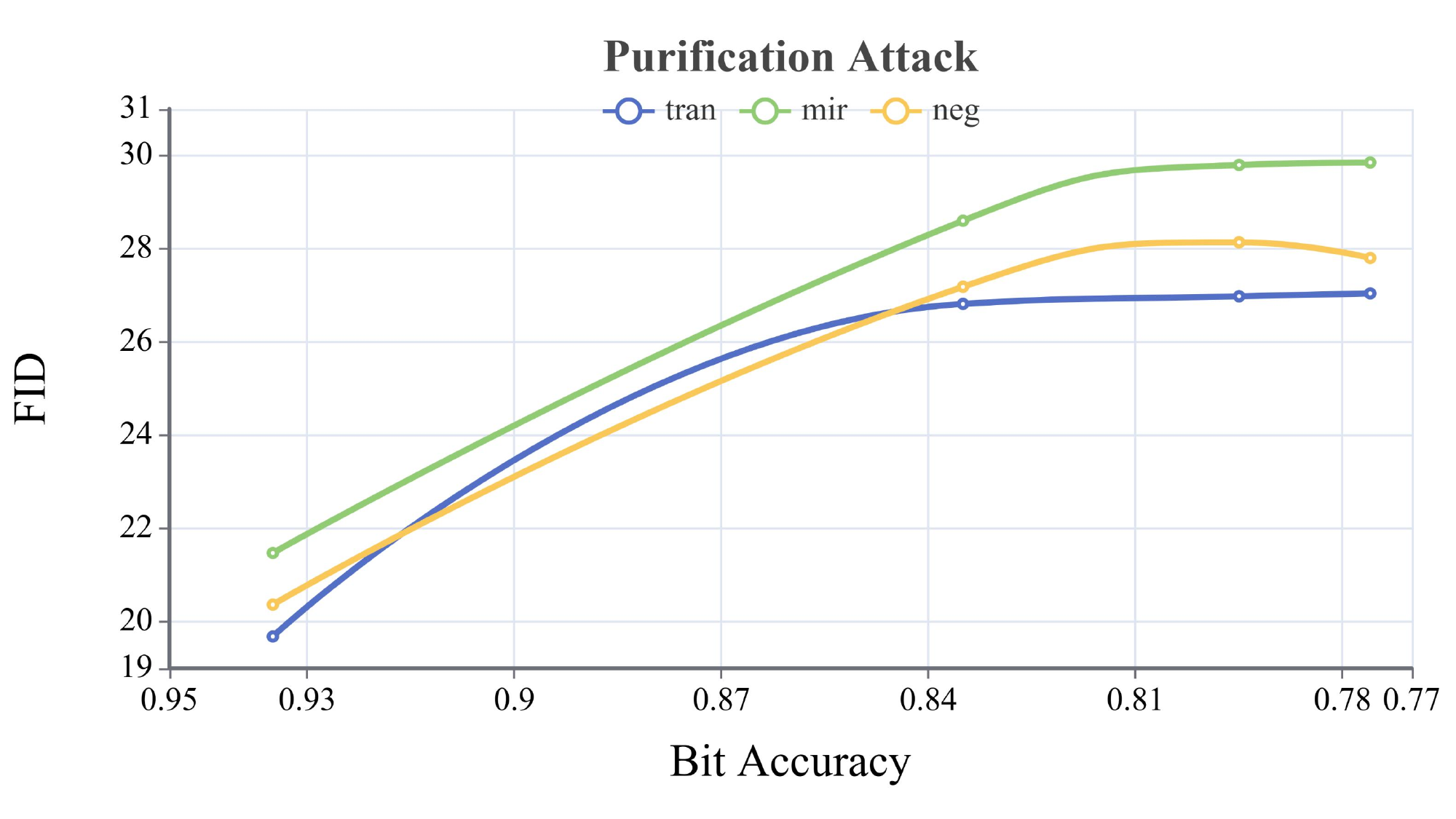}
    \vspace{-2.5em}
    \caption{FID variation during the purification process for 3 different FedOT methods.}
    \vspace{-2em}
    \label{fig:purification_attack}
\end{wrapfigure}
A purification attack targets watermarks or hidden information embedded in model parameters. In this scenario, the attacker fine-tunes or optimizes the model using a clean dataset without any watermark, aiming to "purify" the model by removing the watermark while preserving the model’s original performance as much as possible. In other words, the attacker retrains part of the model with real, unwatermarked data to erase or weaken the previously embedded watermark without significantly degrading the model’s output quality.

We conduct purification attacks by fine-tuning the watermarked parameter components using a clean dataset~\cite{coco}. 
Fig.~\ref{fig:purification_attack} illustrates the change in FID during the watermark purification process. After 300 epochs of purification, the FID of all three LVT methods increased from below 22 to above 26, with $\text{FedOT}_\text{mir}$ showing the largest increase. This demonstrates that removing the watermark inevitably impacts the quality of generated images, validating the robustness of our proposed methods. Empirically, it is very challenging to eliminate watermarks without sacrificing image quality. More robustness evaluations are provided in Appendix~\ref{sec: watermark robustness}.

\begin{table}[t]
\centering
\begin{minipage}{0.48\columnwidth}
\centering
\vspace{0.25em}
\caption{FID, CLIP-Score, Bit Accuracy, and Detection across different numbers of clients.}
\vspace{-1em}
\resizebox{\columnwidth}{!}{
\begin{tabular}{c||c|ccc}
\hline
\rowcolor[HTML]{EFEFEF} 
\cellcolor[HTML]{EFEFEF} & \cellcolor[HTML]{EFEFEF} & \multicolumn{3}{c}{\cellcolor[HTML]{EFEFEF}{\textbf{Number of Clients}}} \\ \cline{3-5} 
\rowcolor[HTML]{EFEFEF} 
\multirow{-2}{*}{\cellcolor[HTML]{EFEFEF}{\textbf{Method}}} & \multirow{-2}{*}{\cellcolor[HTML]{EFEFEF}{\textbf{Metric}}} & 5 & 10 & 20 \\ \hline \hline

\multirow{4}{*}{\shortstack{\textbf{FedOT$_\text{tran}$} \\ \textbf{($c=5$)}}} 
    & FID $\downarrow$         & 19.684 & 20.119 & 19.449 \\
    & CLIP-Score $\uparrow$    & 0.306  & 0.306  & 0.304 \\
    & Bit acc $\uparrow$       & 0.935  & 0.941  & 0.940 \\
    & Detection $\uparrow$     & 0.968  & 0.983  & 0.978 \\
\hline

\multirow{4}{*}{\textbf{FedOT$_\text{mir}$}}  
    & FID $\downarrow$         & 21.475 & 22.540 & 22.22\\
    & CLIP-Score $\uparrow$    & 0.293  & 0.293  & 0.291\\
    & Bit acc $\uparrow$       & 0.926  & 0.928  & 0.926\\
    & Detection $\uparrow$     & 0.962  & 0.970  & 0.967\\
\hline

\multirow{4}{*}{\textbf{FedOT$_\text{neg}$}}  
    & FID $\downarrow$         & 20.367 & 20.541 & 20.23\\
    & CLIP-Score $\uparrow$    & 0.295  & 0.294  & 0.293\\
    & Bit acc $\uparrow$       & 0.922  & 0.914  & 0.910\\
    & Detection $\uparrow$     & 0.953  & 0.947  & 0.941\\
\hline
\end{tabular}
}

\label{tab:different_clients_performance}
\end{minipage}
\hfill
\begin{minipage}{0.48\columnwidth}
\centering
\caption{FID, CLIP-Score, Bit Accuracy, and Detection under different Dirichlet distributions.}
\vspace{-1em}
\resizebox{\columnwidth}{!}{
\begin{tabular}{c||c|ccc}
\hline
\rowcolor[HTML]{EFEFEF} 
\cellcolor[HTML]{EFEFEF} & \cellcolor[HTML]{EFEFEF} & \multicolumn{3}{c}{\cellcolor[HTML]{EFEFEF}{\textbf{Data Distribution}}} \\ \cline{3-5} 
\rowcolor[HTML]{EFEFEF} 
\multirow{-2}{*}{\cellcolor[HTML]{EFEFEF}{\textbf{Method}}} & \multirow{-2}{*}{\cellcolor[HTML]{EFEFEF}{\textbf{Metric}}} & { i.i.d} & {$\alpha = 0.7$} & {$\alpha = 0.5$} \\ \hline \hline

\multirow{4}{*}{\shortstack{\textbf{FedOT$_\text{tran}$} \\ ($c=5$)}}    
    & FID $\downarrow$         & 19.684 & 20.33  & 20.177 \\
    & CLIP-Score $\uparrow$    & 0.306  & 0.306  & 0.306 \\
    & Bit acc $\uparrow$       & 0.935  & 0.942  & 0.941 \\
    & Detection $\uparrow$     & 0.968  & 0.983  & 0.980 \\
\hline

\multirow{4}{*}{\textbf{FedOT$_\text{mir}$}}  
    & FID $\downarrow$         & 21.475 & 22.294 & 22.437 \\
    & CLIP-Score $\uparrow$    & 0.293  & 0.290  & 0.289 \\
    & Bit acc $\uparrow$       & 0.926  & 0.935  & 0.935 \\
    & Detection $\uparrow$     & 0.962  & 0.974  & 0.976 \\
\hline

\multirow{4}{*}{\textbf{FedOT$_\text{neg}$}}  
    & FID $\downarrow$         & 20.367 & 21.491 & 21.380 \\
    & CLIP-Score $\uparrow$    & 0.295  & 0.293  & 0.293 \\
    & Bit acc $\uparrow$       & 0.922  & 0.924  & 0.921 \\
    & Detection $\uparrow$     & 0.953  & 0.960  & 0.957 \\

\hline
\end{tabular}
}
\label{tab:non_iid_distribution}
\end{minipage}
\end{table}

\section{Conclusion}

This paper introduces \textbf{FedOT}, the first ownership verification and tracing framework for Latent Diffusion Models (LDMs) in Federated Learning. 
To identify the responsible client for a leak, we propose a chunked watermark for ownership verification and traceability. 
The chunked watermark addresses a critical gap in protecting and ensuring traceability for generative models within FL.
Unlike existing VAE-based watermarking methods, which are vulnerable to removal attacks, FedOT leverages \textbf{Latent Vector Transformation (LVT)} to modify the latent space of the VAE, effectively binding the VAE and U-Net. Ensuring that any attempt to replace the VAE for watermark removal leads to severe degradation in image quality.
We propose three LVT strategies, including translation, mirror, and negative transformation.
Although the replacement attack can remove the watermark, it comes at the cost of a severe loss in image quality. This trade-off renders the attack impractical and demonstrates the robustness of our method.

\clearpage  

\section*{Acknowledgements}
This work is supported by the National Natural Science Foundation of China (No. 62436007, No. 62572147)

%
%

\bibliographystyle{splncs04}
\bibliography{main}

\clearpage
\appendix
\section*{\huge Appendix}

\startcontents[supp]

\makeatletter
\renewcommand{\theHsection}{supp.\Alph{section}}
\renewcommand{\theHsubsection}{supp.\Alph{section}.\arabic{subsection}}
\makeatother

\setcounter{figure}{5}
\setcounter{table}{7}
\setcounter{equation}{8}
\setcounter{algorithm}{1}

\renewcommand{\thefigure}{\arabic{figure}}
\renewcommand{\thetable}{\arabic{table}}
\renewcommand{\theequation}{\arabic{equation}}
\renewcommand{\thealgorithm}{\arabic{algorithm}}

\newcommand{\suptocA}[2]{%
  \noindent\hyperref[#1]{\textbf{#2}}\dotfill\pageref{#1}\par\vspace{2pt}%
}
\newcommand{\suptocB}[2]{%
  \noindent\hspace{1.5em}\hyperref[#1]{#2}\dotfill\pageref{#1}\par\vspace{2pt}%
}

\section*{Table of Contents}
\suptocA{sec:aa}{A\quad Preliminaries}
\suptocB{sec:A1}{A.1\quad Federated LDMs and Threat Model}
\suptocB{sec:A2}{A.2\quad Local Training under FedOT}
\suptocA{sec:Watermark_Design_and_Training_Details}{B\quad Watermark Design and Training Details}
\suptocB{sub_sec:More_watermark_design_details}{B.1\quad Additional Watermark Design Details}
\suptocB{sub_sec:More_watermark_training_details.}{B.2\quad Additional Watermark Training Details}
\suptocB{sub_sec:Bit_Accuracy_and_Detection.}{B.3\quad Definitions of Bit Accuracy and Detection}
\suptocA{sec:CC}{C\quad More Details on Fine-tuning VAE with LVT}
\suptocB{sub_sce:lvt_finetune}{C.1\quad Fine-tuning Details}
\suptocB{sub_sec:LVT_analyze}{C.2\quad Analysis of LVT}
\suptocA{sec: watermark robustness}{D\quad Watermark Robustness}
\suptocB{sub_sec:Generated_Image_Attacks.}{D.1\quad Generated Image Attacks}
\suptocB{sub_sec:Negative_Recovered_Attack}{D.2\quad Negative Recovered Attack}
\suptocB{sub_sec:Translation_Recovered_Attack}{D.3\quad Translation Recovered Attack}
\suptocB{sub_sec:Mirror_Recovered_Attack}{D.4\quad Mirror Recovered Attack}
\suptocB{sub_sec:collusion_attack}{D.5\quad Collusion Attack}
\suptocA{sec:more_experimental_results}{E\quad More Experimental Results}
\suptocB{sub_sec:Flicker30K_Dataset}{E.1\quad Results on Flicker30K Dataset}
\suptocB{sub_sec:VAE_image_reconstruction_quality}{E.2\quad VAE Image Reconstruction Quality}
\suptocB{sub_sec:watermark_image_generation_quality}{E.3\quad Image Generation Quality}
\suptocB{sub_sec:tau_detection}{E.4\quad Selection of Detection Threshold $\tau$}
\suptocB{sec:end-to-end}{E.5\quad End-to-End Attribution Accuracy}
\suptocB{sub_sec:More_FedOT_visual_results}{E.6\quad Additional Visualizations}
\suptocA{sub_sec:Secrecy_of_the_LVT}{F\quad Additional Discussions}
\suptocB{sub_sec:Secrecy_of_the_LVT}{F.1\quad Secrecy of LVT}
\suptocB{sub_sec:limitation}{F.2\quad Limitations}
\suptocB{sub_sec:Communication_overhead}{F.3\quad Communication Overhead}
\suptocB{sub_sec:Scalability}{F.4\quad Scalability}

\section{Preliminaries}
\label{sec:aa}
\subsection{Federated LDMs and Threat Model}
\label{sec:A1}
\noindent\textbf{Federated Latent Diffusion Models.}
Fig.~{1}(a) in the main paper illustrates the overall structure of Federated Latent Diffusion Models (FedLDMs).
A global Latent Diffusion Model (LDM) is maintained on a central server, which contains:
\begin{itemize}
\item A pretrained VAE~\cite{vae} that encodes images into a latent space, and

\item A diffusion model operating in the latent space using a U-Net architecture~\cite{u-net}.
\end{itemize}
During federated training, each client updates only the U-Net parameters while keeping the VAE frozen. Malicious clients may store intermediate global models and redistribute them illegally.

\vspace{0.3em}\noindent\textbf{Watermarking for Ownership Verification and Traceability.}
To prevent unauthorized model leakage, an $n$-bit chunked watermark $\mathbf{m}$ is embedded into each distributed LDM.
The embedded watermark enables:
\begin{itemize}
\item Ownership verification — confirming whether a given image originates from the federated LDM.

\item Traceability — determining which client is responsible for a leaked model.
\end{itemize}
The watermark embedding procedure is shown in Appendix~\ref{sec:Watermark_Design_and_Training_Details}
.

\vspace{0.3em}\noindent\textbf{Threat Model.}
In the federated learning setup, the server and clients operate under different assumptions:
\begin{itemize}
\item The server is trusted and embeds unique watermarks into distributed models, aggregates client updates, and performs watermark-based verification when suspicious images appear.

\item The client group $C=\{c_i\}_{i=1}^K$ may contain malicious clients. These clients may attempt to evade watermark detection through image manipulation, model fine-tuning, or direct modification of watermark-related parameters.
\end{itemize}

\subsection{Local Training under FedOT}
\label{sec:A2}
The overall federated training process of Stable Diffusion under the proposed FedOT framework is shown in Algorithm~\ref{alg:federated_sd}.
Given the watermarked models $\{\hat{M}_i\}_{i=1}^K$, the total training epochs $S$, the number of clients $K$, and the client datasets $\{D_i\}_{i=1}^K$, each client receives its own watermarked model and performs local training.
During each local update, only the U-Net parameters are optimized, while the VAE and text encoder remain frozen to preserve watermark integrity.
After all clients finish their local updates, the server aggregates the U-Net parameters and redistributes the aggregated model for the next round.
Repeating this process yields the final global model $M_g^T$.

\begin{figure}[t]
\centering
\begin{minipage}{0.48\columnwidth}
    \centering
    \includegraphics[width=\linewidth]{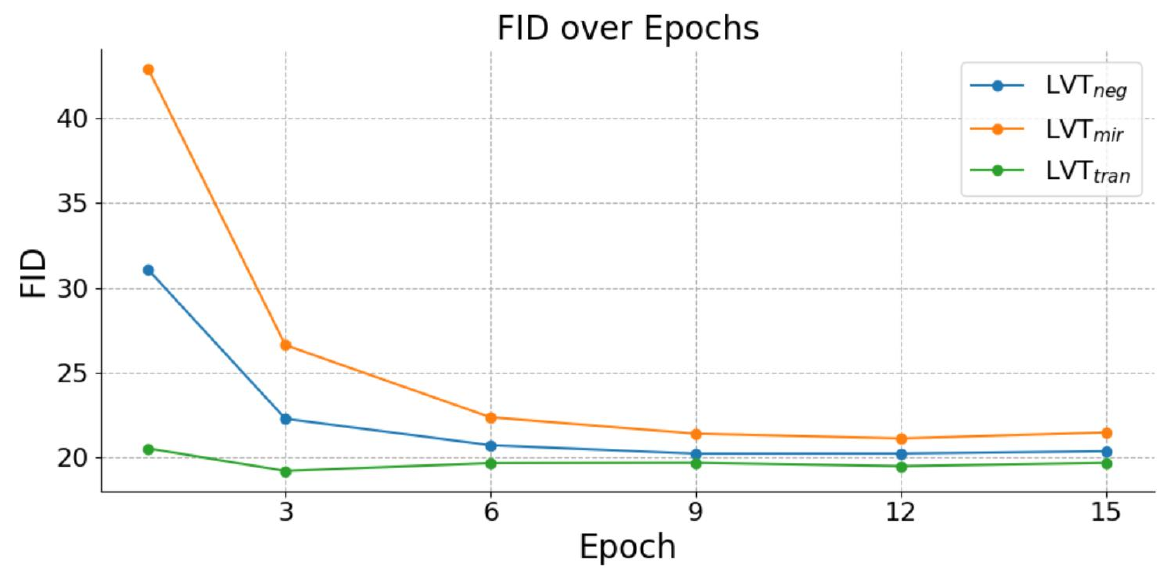}
    \caption{FID trends across training rounds under FedOT framework.}
    \label{fig:fid_over_epoch}
\end{minipage}
\hfill
\begin{minipage}{0.48\columnwidth}
    \centering
    \includegraphics[width=\linewidth]{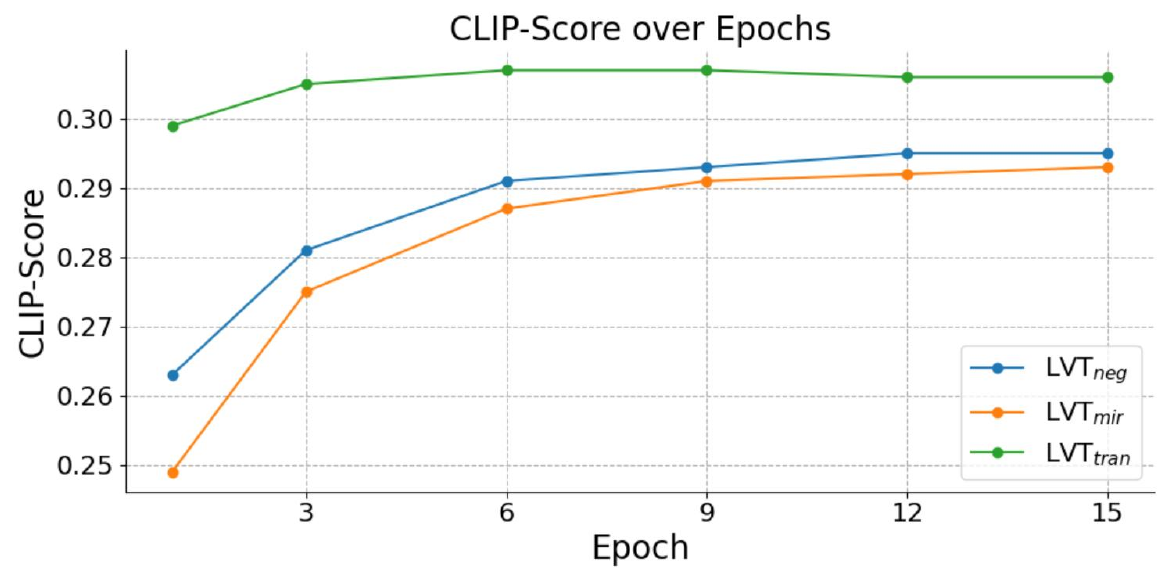}
    \caption{CLIP-Score trends across training rounds under FedOT framework.}
    \label{fig:clip_over_epoch}
\end{minipage}
\end{figure}

Unlike typical generative model training, where performance improves steadily with more epochs, the FedOT framework exhibits different learning dynamics due to the LVT-induced transformation of the latent space.
Through extensive experiments, we observe that the generated image quality first improves and then degrades as federated training progresses.
As shown in Figs.~\ref{fig:fid_over_epoch} and \ref{fig:clip_over_epoch}, both FID and CLIP-Score stabilize around epoch = 15, after which additional training leads to deterioration in image quality.
Based on these observations, we adopt 15 federated epochs for all experiments in this paper.

\begin{algorithm}[t]
\caption{Federated Training SD in FedOT}
\label{alg:federated_sd}
\begin{algorithmic}[1]
\item[\textbf{Input:}] Watermarked models $\{\hat{M}_k\}_{k=1}^K$, training epochs $S$, number of clients $K$, clients' datasets $\{D_i\}_{i=1}^K$
\item[\textbf{Output:}] Aggregated global model $M_g^T$

\For{$i \gets 1$ \textbf{to} $K$}
    \State $\hat{M}_i^0 \gets \hat{M}_i$ \hspace{2em}\graycomment{Distribute complete models in the first round}
\EndFor

\For{$t \gets 1$ \textbf{to} $S$}
    \For{$i \gets 1$ \textbf{to} $K$}
        \State $\hat{M}_i^t \gets LocalTrain(\hat{M}_i^{t-1}, D_i)$
    \EndFor
    \State $\hat{M}_g^t \gets \sum_{i=1}^{K} \frac{|D_i|}{\sum_{j=1}^{K} |D_j|} \hat{M}_i^t$\hspace{2em}  \graycomment{Aggregate only U-Net parameters}
    \If{$t < S$}
        \For{$i \gets 1$ \textbf{to} $K$}
            \State $\hat{M}_i^t \gets \hat{M}_g^t$ \hspace{2em}\graycomment{Update client U-Net}
        \EndFor
    \EndIf
\EndFor
\State $M_g^T = \hat{M}_g^t$ \hspace{2em}\graycomment{Upload the complete model in the final round}
\State \Return $M_g^T$
\end{algorithmic}
\end{algorithm}

\begin{algorithm}[t]
\caption{Watermark Embedding in FedOT}
\label{alg:embed_watermark}
\begin{algorithmic}[1]
\Statex \hspace*{-1.2em}\textbf{Input:} Global model $M_T$, number of clients $K$ , length of watermark $n$, length of verification $r$ 
\Statex \hspace*{-1.2em}\textbf{Output:} Watermarked models $\{\hat{M}_i\}_{i=1}^K$

\For{$i \gets 1$ \textbf{to} $K$}
    \State $M_i \gets CopyModel(M_T)$ 
\EndFor

\State $m_{1:r} \gets RandomBitString(r)$ 

\For{$i \gets 1$ \textbf{to} $K$}
    \State $m_{i,r+1:n} \gets HD(RandomBitString(n-r),K)$ 
\EndFor

\For{$i \gets 1$ \textbf{to} $K$}
    \State $m_i \gets Concat(m_{1:r}, m_{i,r+1:n})$ 
    \State $\hat{M}_i \gets EmbedWatermark(M_i, m_i)$
\EndFor

\State \Return $\{\hat{M}_i\}_{i=1}^K$
\end{algorithmic}
\end{algorithm}

\section{Watermark Design and Training Details}
\label{sec:Watermark_Design_and_Training_Details}
In this section, we provide additional implementation details that are not explicitly described in the main text.
In FedOT, the watermark embedding process takes as input the global model $M$, the number of clients $K$, the watermark length $n$, and the ownership verification length $r$.
First, the global model is replicated $K$ times to obtain the model copies $\{M_i\}_{i=1}^K$.
Next, the first $r$ bits of the watermark are generated for ownership verification.
Then, following the Hamming Distance optimization strategy, the remaining $n-r$ bits of the watermark are generated for each client to ensure sufficient inter-client separability.
Finally, the first $r$ bits and the optimized $n-r$ bits are concatenated to form $K$ complete $n$-bit watermarks, which are subsequently embedded into their corresponding model copies, resulting in the watermarked models $\{\hat{M}_i\}_{i=1}^K$.
The detailed procedure is shown in Algorithm~\ref{alg:embed_watermark}.

\subsection{Additional Watermark Design Details}
\label{sub_sec:More_watermark_design_details}
To minimize the risk of confusion, which refers to misidentifying the source client due to similar tracing bits, we draw inspiration from prior work~\cite{fedtracker} and optimize their Hamming Distance (HD) between the suffixes of different client watermarks. Specifically, we define the watermark set  $\{\textbf{m}_i\}_{i=1}^{K}$ with the following objective:
\begin{equation}
\{\textbf{m}_i\}_{i=1}^{K} = \arg \max \min_{1 \leq i < j \leq K} HD(\textbf{m}_{i,r+1:n}, \textbf{m}_{j,r+1:n}),
\end{equation}
we use a Genetic Algorithm (GA)~\cite{genetic} for approximate optimization, maximizing watermark distinction across clients. To further validate the scalability of our tracing mechanism beyond the experimental scale evaluated in the main paper, we conduct large-scale client simulations. As shown in Table~\ref{tab:collision_supp}, random assignment yields a collision probability of 0.303\% at $10^2$ clients, while our Hamming-optimized assignment maintains 0\% collision probability up to $10^3$ clients, demonstrating reliable tracing capability well beyond the client scale used in our main experiments.

\begin{table}[t]
\centering
\caption{Collision probability under random vs. Hamming-optimized client ID assignment across varying client scales.}
\vspace{-1em}
\small
\resizebox{0.6\columnwidth}{!}{
\begin{tabular}{c|c||ccccc}
\hline
\rowcolor[HTML]{EFEFEF} 
\cellcolor[HTML]{EFEFEF} & \cellcolor[HTML]{EFEFEF} & \multicolumn{5}{c}{\cellcolor[HTML]{EFEFEF}{\textbf{Number of Clients}}} \\
\cline{3-7}
\rowcolor[HTML]{EFEFEF} 
\multirow{-2}{*}{\cellcolor[HTML]{EFEFEF}{\textbf{Method}}} & \multirow{-2}{*}{\cellcolor[HTML]{EFEFEF}{\textbf{Metric}}} & 10 & $10^2$ & $10^3$ & $10^4$ & $10^5$ \\ \hline \hline
Random & \multirow{2}{*}{Collision prob.} & 0\% & 0.303\% & 0.343\% & 0.349\% & 0.350\% \\
Hamming-opt. & & 0\% & 0\% & 0\% & 0.337\% & 0.349\% \\
\hline
\end{tabular}
}
\vspace{-0.5em}
\label{tab:collision_supp}
\end{table}

\subsection{Additional Watermark Training Details}
\label{sub_sec:More_watermark_training_details.}
To embed a unique watermark into each client model on the server side, we adopt the watermark extractor $E$ from Stable Signature~\cite{Stable_signature} during training. Specifically, we first generate a unique binary watermark $\textbf{m}$ for each client. A set of public images is then sampled and passed through a frozen VAE encoder to obtain latent representations $z$. These latent vectors use a trainable VAE decoder to reconstruct images $\hat{x}$, which are then fed into the extractor $E$ to predict the embedded watermark $\textbf{m}'$. Importantly, the VAE encoder is kept frozen throughout this process to preserve the latent distribution, ensuring that downstream generation quality is not compromised.

\begin{figure}[t]
\centering
\includegraphics[width=\columnwidth]{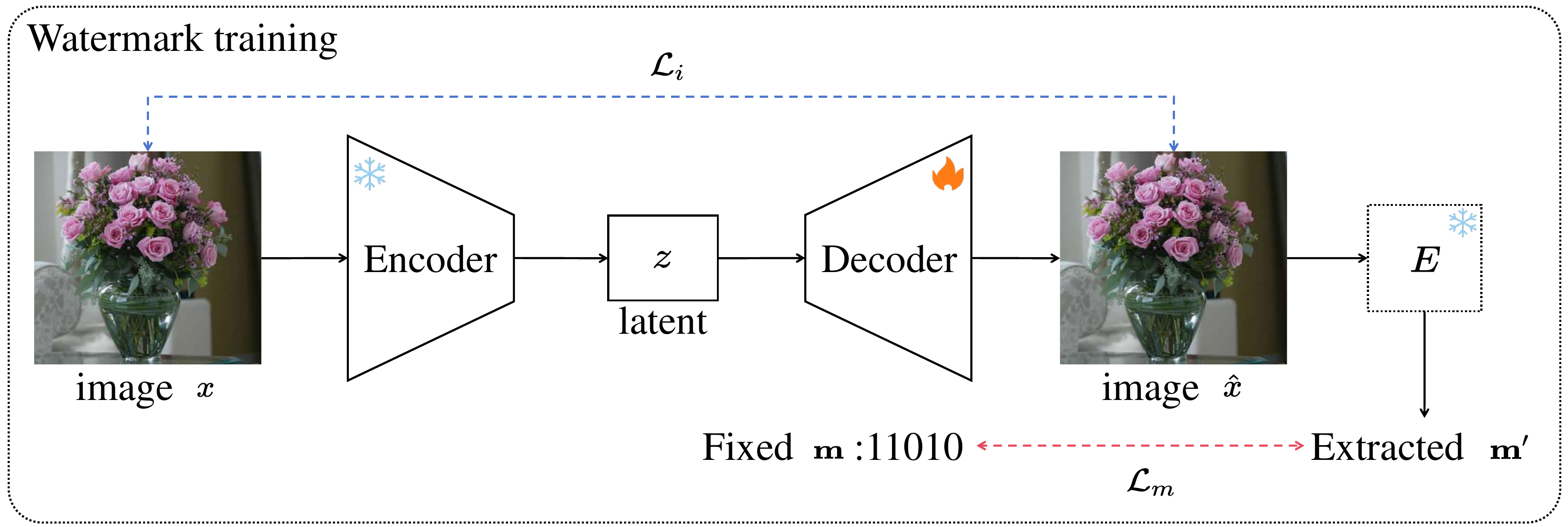}
\vspace{-2em}
\caption{Training a watermark in decoder with extractor $E$.}
\vspace{-1.25em}
\label{fig:training_watermark}
\end{figure}

As illustrated in Fig.~\ref{fig:training_watermark}, this procedure allows the server to produce a watermarked decoder for each client. These decoders are capable of invisibly embedding client-specific watermarks into generated outputs, enabling ownership verification without affecting visual fidelity or model utility. 

\begin{algorithm}[t]
\caption{LVT Training in FedOT}
\label{alg:LVT_training}
\begin{algorithmic}[1]
\Statex \hspace*{-1.2em}\textbf{Input:} Global Model $M$, transformation $T$, datasets $Data$, training steps $S$, constant $c$
\Statex \hspace*{-1.2em}\textbf{Output:} LVT global model $M_T$
\State $E, D \gets GetVAEEncoderDecoder(M)$
\Statex\graycomment{Stage I: Train Encoder $E_T$}
\For{$i \gets s$ \textbf{to} $S$}
    \State $x \gets ImageProcessing(Data,batchsize)$
    \State $z = E(x)$
    \If{$T = \text{translation}$}
        \State $z' \gets z + c$
    \ElsIf{$T = \text{mirror}$}
        \State $z' \gets -z$
    \Else
        \State $z' \gets z$
        \State $x \gets 255-x$
    \EndIf
    \State $x'=D(z')$
    \State $\mathcal{L}_{\text{VAE}}(x,x')$ \hspace{2em}\graycomment{Update $E$ by minimizing $\mathcal{L}_{\text{VAE}}$}
\EndFor
\State $E_T \gets E$
\Statex\graycomment{Stage II: Train Decoder $D_T$}
\For{$i \gets s$ \textbf{to} $S$}
    \State $x \gets ImageProcessing(Data,batchsize)$
    \State $z^* \gets E_T(x)$
    \State $x^*\gets D(z^*)$
    \State $\mathcal{L}_{\text{VAE}}(x,x^*)$ \hspace{2em}\graycomment{Update $D$ by minimizing $\mathcal{L}_{\text{VAE}}$}
\EndFor
\State $D_T \gets D$
\State $M_T \gets UpdateGlobalModel(M,E_T,D_T)$
\State \Return $M_T$
\end{algorithmic}
\end{algorithm}

\subsection{Definitions of Bit Accuracy and Detection}
\label{sub_sec:Bit_Accuracy_and_Detection.}
\noindent\textbf{Bit Accuracy (Bit Acc).}  
Bit accuracy quantifies the proportion of correctly extracted bits in the predicted watermark $\textbf{m}'$ compared to the ground-truth $\textbf{m}$, and is defined as:
\begin{equation}
\text{Match}(\textbf{m}, \textbf{m}') = \frac{1}{n} \sum_{i=1}^{n} (\textbf{m}[i] = \textbf{m}'[i]).
\end{equation}
This is equivalent to the Bit Accuracy (Bit Acc).

\vspace{0.3em}\noindent\textbf{Detection.}  
Detection measures the proportion of watermarked images whose extracted watermark achieves a bit accuracy above a predefined threshold $\tau$. Formally, given a set of $N$ generated images with corresponding extracted watermarks $\textbf{m}'$ and ground-truth watermarks $\textbf{m}$, the detection rate is computed as:
\begin{equation}
\text{Detection} = \frac{1}{N} \sum_{}^{} \mathbb{I} \left(\text{Match}(\textbf{m}, \textbf{m}') \geq \tau \right),
\end{equation}
where $\mathbb{I}(\cdot)$ denotes the indicator function that equals 1 when the condition inside holds true, and 0 otherwise. A higher detection rate indicates better robustness and reliability of watermark extraction.

\section{More Details on Fine-tuning VAE with LVT}
\label{sec:CC}
\subsection{Fine-tuning Details}
\label{sub_sce:lvt_finetune}
In the Methodology section, we described the training process of adapting the VAE to Latent Vector Transformation (LVT). Here, we provide additional implementation details and clarification.

During the entire VAE fine-tuning process, the encoder learns the forward transformation $T$, while the decoder learns its inverse transformation $T^{-1}$. This adjustment modifies the latent vectors passed to the U-Net, allowing them to gradually adapt to the new distribution during the diffusion process, thereby establishing a strong coupling between the U-Net and the VAE. 
In the translation transformation, we set a constant $c = 5$ as the translation parameter.
In the mirror transformation, we perform mirroring by multiplying the mean $\mu$ by $-1$.
For the negative transformation, we use the negative version of the image as the ground truth, enabling the VAE encoder to learn a pixel-wise negative mapping.
After the encoder learns each corresponding transformation, the decoder is trained to adapt to the new latent space and reconstruct the images accordingly.
The overall training process of the LVT is shown in Algorithm~\ref{alg:LVT_training}.

The optimization is performed using the following loss functions. Mean Squared Error (MSE) Loss: Ensures that the reconstructed image $x'$ remains close to the original image $x$, which is defined as \( \mathcal{L}_\text{MSE} = \| x - x' \|^2 \).
Additionally, the perceptual loss~\cite{perceptual} captures high-level perceptual features by comparing the outputs of a pre-trained model applied to the original and reconstructed images and is defined as:
\begin{equation}
\mathcal{L}_\text{p} = \frac{1}{n} \sum_{i=1}^{n} \|\phi_i(x) - \phi_i(x')\|_2^2,
\end{equation}
where $\phi_i$ represents the $i$-th feature map of VGG19.

Kullback-Leibler (KL) Divergence~\cite{kl-div} regularizes the latent space distribution to maintain smoothness and prevent overfitting, which is defined as \( D_\text{KL}(q(z|x) \allowbreak\, || \, p(z)) \), where \( q(z|x) \) represents the distribution of \( z \) given \( x \), and \( p(z) \) is the standard normal distribution \( \mathcal{N}(0, I) \).The KL loss can be further expressed as:
\begin{equation}
\mathcal{L}_\text{KL} = \frac{1}{2} \sum_{i} \left( \mu_i^2 + \sigma_i^2 - 1 - \log \sigma_i^2 \right).
\end{equation}

The KL divergence is weighted by a small factor \(\lambda_\text{KL} = 10^{-8}\) to control its influence on the overall loss.

Generator Loss~\cite{taming} encourages the decoder to produce realistic images that can fool a discriminator, improving the quality of generated images:
\begin{equation}
\mathcal{L}_\text{gen} = -\mathbb{E}[\text{discriminator}(x’)].
\end{equation}

To balance the influence of the perceptual and generator losses, we calculate an adaptive weight:
\begin{equation}
\lambda_\text{ada} = \frac{\|\nabla_{\theta_\text{dec}} \mathcal{L}_\text{p}\|_2}{\|\nabla_{\theta_\text{dec}} \mathcal{L}_\text{gen}\|_2 + \delta},
\end{equation}
where $\theta_\text{dec}$ represents the parameters of the VAE decoder, and $\delta$ is a small constant to prevent division by zero. The adaptive weight is further clamped to a maximum value of $10^4$ to maintain training stability.

The overall VAE training loss with LVT integrates all components as follows:
\begin{equation}
\mathcal{L}_{\text{VAE}} = \mathcal{L}_{\text{MSE}} + \mathcal{L}_\text{p} + \lambda_{\text{KL}} \cdot \mathcal{L}_{\text{KL}} + \lambda_{\text{ada}} \cdot \mathcal{L}_{\text{gen}}.
\end{equation}
This composite objective ensures faithful reconstruction, perceptual similarity, latent space regularization, and generation quality, enabling the VAE to effectively adapt to the transformed latent space.

\subsection{Analysis of LVT}
\label{sub_sec:LVT_analyze}
We denote the latent vector sampled from the encoder as:
\begin{equation}
z = \mu + \sigma \cdot \epsilon, \quad \epsilon \sim \mathcal{N}(0, I),
\end{equation}
here, $\epsilon$ is the reparameterization noise used for sampling from the latent distribution of encoder, and $z \sim \mathcal{N}(\mu, \sigma^2).$
During LVT, a transformation $T(\cdot)$ is applied to $z$, yielding:
\begin{equation}
z' = T(z).
\end{equation}
The reconstructed image is $x'= D(z')$, where $D$ denotes the decoder.
The reconstruction error depends on how $T(\cdot)$ changes the distribution and local topology of the latent space.

\vspace{0.3em}\noindent\textbf{Random Transformation.} 
Directly adding random Gaussian noise to the latent vectors in LVT significantly degrades image reconstruction quality. For a random perturbation:
\begin{equation}
z'=z + \hat\epsilon, \quad \hat\epsilon\sim \mathcal{N}(0 ,I),
\end{equation}
where $\hat{\epsilon}$ denotes an independently sampled perturbation noise added to the latent vector. We obtain:
\begin{equation}
z'\sim \mathcal{N}(\mu ,I(\sigma^2+1)).
\end{equation}
While the injected noise increases the variance to $\sigma^2 + 1$. Although such a distributional shift can theoretically be learned, the original latent vectors are continuous, and the random noise added at each training iteration disrupts the local continuity of the latent space, for any pair of neighboring samples $z_i, z_j$ in the latent vector $z$, we have:
\begin{equation}
\| (z_i + \epsilon_i) - (z_j + \epsilon_j) \|_2^2 = \| z_i - z_j \|_2^2 + \| \epsilon_i - \epsilon_j \|_2^2.
\end{equation}
The latter term is dominated by noise, which disrupts the local neighborhood structure of the latent space, making it difficult for the decoder $D$ to learn a stable inverse mapping.
Consequently, the decoder fails to accurately reconstruct the images, leading to a noticeable degradation in overall reconstruction quality.

\vspace{0.3em}\noindent\textbf{Translation and Mirror Transformations.} 
Both translation and mirror transformations are deterministic and globally consistent linear mappings that can be expressed as:
\begin{equation}
z' = T(z) = A z + C,
\end{equation}
where for translation, $A = 1$ and $C \neq 0$, and for mirroring, $A = -1$ and $C = 0$. 
Under such transformations, the latent distribution becomes:
\begin{equation}
z' \sim \mathcal{N}(A\mu + C, \sigma^2).
\end{equation}
For any pair of neighboring latent vectors $z_i, z_j$, we have:
\begin{equation}
\| z_i' - z_j' \| = \| A(z_i - z_j) \| = \| z_i - z_j \|,
\end{equation}
indicating that the local geometric structure of the latent space is preserved. 

Compared to adding random Gaussian noise, these structured transformations introduce a global and consistent shift without altering the variance of the distribution. 
This enables the decoder to perform stable reconstruction while embedding a globally detectable perturbation within the latent space. 
The preservation of local relationships among latent vectors ensures that the generated images maintain high visual quality despite the applied transformations.

\vspace{0.3em}\noindent\textbf{Negative Transformations.} 
The effectiveness of the negative transformation lies in the fact that it preserves the overall distribution of the latent space while introducing a structured, pixel-wise mapping. Each latent dimension encodes a consistent negative relationship, maintaining the Gaussian property yet producing a globally detectable transformation. During training, the encoder learns to map inputs into the latent representations of their negative counterparts, while the decoder adapts accordingly to reconstruct images from these transformed representations.

\section{Watermark Robustness}
\label{sec: watermark robustness}
\subsection{Generated Image Attacks}
\label{sub_sec:Generated_Image_Attacks.}
To evaluate the robustness of the embedded watermark against post-processing, we apply a variety of common image-level attacks to the generated images. These include basic geometric and photometric transformations such as cropping (Crop), brightness adjustment (Brigh.), JPEG compression (JPEG50), contrast adjustment (Cont), text overlay, and resizing to 50\% of the original resolution (Resize 0.5). These perturbations simulate real-world image editing scenarios that may occur during content sharing or malicious tampering.

As shown in Table~\ref{tab:com_image_attack}, we evaluated the robustness of the watermark in generated images under various image-level attacks.  Results show that although the accuracy of our proposed FedOT decreases slightly compared to Stable Signature* under these perturbations, it still demonstrates strong overall performance. 

In particular, under the “Comb.” attack combining Crop and Brigh., the worst-case bit accuracy remains as high as 0.772, demonstrating the robustness of our method.

\subsection{Negative Recovered Attack}
\label{sub_sec:Negative_Recovered_Attack}
In our design of LVT, the introduced negative transformation causes the generated images to exhibit a distinct negative-like appearance when a malicious client attempts to remove the watermark by replacing the VAE. This phenomenon makes it easy for the attacker to realize that the degradation in image quality is due to the latent space having learned a pixel-level negative mapping. To mitigate this degradation, the attacker may apply an additional negative transformation to the generated images in an attempt to restore them to normal appearance. To account for this potential countermeasure, we conducted a simulated experiment to evaluate the effectiveness of this “double negative recovery” strategy. 

As shown in Table~\ref{tab:neg_attack}, Clean denotes the generation results with watermarking but without any attack, Attack refers to the results after replacing the VAE to remove the watermark, and Recovered represents the results obtained by attackers who, upon observing the negative-like artifacts, attempt to restore the image by applying an additional negative transformation. The results show that even after this recovery operation, the FID score remains high, indicating that the watermark still leads to noticeable degradation in image quality.

This phenomenon can be attributed to the fact that the negative mapping learned in the latent space is only an approximation. Due to constraints from the model architecture and optimization objectives, the latent space cannot perfectly replicate a pixel-level negative transformation. As a result, the attacker’s recovery via an additional negative operation can only partially reverse the effect. The restored images still deviate from the original ones, leading to consistently high FID scores.

\begin{table*}[t]
\centering
\caption{presents the impact of different image attacks on watermark accuracy under the three LVT methods.}
\vspace{-1em}
\scriptsize
\arrayrulewidth=0.08mm
\resizebox{\textwidth}{!}{
  \begin{tabular}{c||ccccccc}
    \hline
    \rowcolor[HTML]{EFEFEF} 
    \textbf{Method} & Crop $\uparrow$ & Brigh. $\uparrow$ & JPEG50 $\uparrow$ & Cont $\uparrow$ & Text overlay $\uparrow$ & Resize 0.5 $\uparrow$ & Comb. \textbf{} \\ 
    \hline \hline
    Stable Signature*  & 0.952 & 0.934 & 0.865 & 0.931 & 0.966 & 0.873 & 0.853   \\
    $\text{FedOT}_\text{tran}$ & {0.912} & {0.884} & {0.813} & {0.877} & {0.920} & {0.803} & {0.796} \\ 
    $\text{FedOT}_\text{mir}$ & 0.897 & 0.868 & 0.800 & 0.855 & 0.909 & 0.778 & 0.772 \\ 
    $\text{FedOT}_\text{neg}$ & 0.897 & 0.865 & 0.804 & 0.861 & 0.904 & 0.779 & 0.778 \\ 
    \hline
  \end{tabular}
  }
\vspace{-1.25em}
\label{tab:com_image_attack}
\end{table*}

\subsection{Translation Recovered Attack}
\label{sub_sec:Translation_Recovered_Attack}
In our LVT design, the translation transformation causes noticeable color shifts when a malicious client attempts to remove the watermark by replacing the VAE. As a result, the attacker can easily perceive the degradation in image quality and infer that the latent space has learned a translation mapping. To restore the visual quality, the attacker may further apply an additional translation transformation to the latent space in an attempt to compensate for the shift. 

To evaluate this scenario, we simulate the most favorable case for the attacker, where the translation parameter is fully known and used to recover the latent representations of the VAE. As shown in Table~\ref{tab:tran_attack}, even when the attacker has complete knowledge of the translation coefficient = 5, the recovered image quality remains low, and in some cases, performs worse than directly replacing the VAE to evade watermarking. This result demonstrates that merely knowing the translation parameter is insufficient to restore high-quality generation, further verifying the robustness of LVT against replacement attacks.

\begin{table*}[t]
\centering
\begin{minipage}{0.48\textwidth}
    \centering
    \caption{Evaluation of $\text{FedOT}_\text{neg}$ under recovered attacks.}
    \vspace{-1em}
    \resizebox{\linewidth}{!}{
      \begin{tabular}{c||cccc}
        \hline
        \rowcolor[HTML]{EFEFEF} 
        \textbf{$\text{FedOT}_\text{neg}$} & FID & CLIP-Score & Detection & Bit Acc \\ 
        \hline \hline
        Clean  & 20.367  & 0.295  & 0.916 & 0.922 \\ 
        Attack & 40.537 \textcolor{green}{(+20.170)}  & 0.272 \textcolor{red}{(-0.23)}  & 0.000 & 0.524 \\ 
        Recovered & 25.069 \textcolor{green}{(+4.702)}  & 0.295  & 0.000 & 0.519 \\ 
        \hline
      \end{tabular}
    }
    \label{tab:neg_attack}
\end{minipage}
\hfill
\begin{minipage}{0.48\textwidth}
    \centering
    \caption{Evaluation of $\text{FedOT}_\text{tran}$ under recovered attacks.}
    \vspace{-1em}
    \resizebox{\linewidth}{!}{
      \begin{tabular}{c||cccc}
        \hline
        \rowcolor[HTML]{EFEFEF} 
        \textbf{$\text{FedOT}_\text{tran}$} & FID & CLIP-Score & Detection & Bit Acc \\ 
        \hline \hline
        Clean  & 19.648  & 0.306  & 0.940 & 0.935 \\ 
        Attack & 20.749 \textcolor{green}{(+1.065)}  & 0.305 \textcolor{red}{(-0.01)}  & 0.000 & 0.513 \\ 
        Recovered & 22.841 \textcolor{green}{(+3.157)}  & 0.303 \textcolor{red}{(-0.03)}  & 0.000 & 0.503 \\ 
        \hline
      \end{tabular}
    }
    \label{tab:tran_attack}
\end{minipage}
\vspace{-0.5em}
\end{table*}

\begin{table}[t]
\centering
\caption{Evaluation of $\text{FedOT}_\text{mir}$ under recovered attacks.}
\vspace{-1em}
\resizebox{0.48\columnwidth}{!}{
  \begin{tabular}{c||cccc}
    \hline
    \rowcolor[HTML]{EFEFEF} 
    \textbf{$\text{FedOT}_\text{mir}$} & FID & CLIP-Score & Detection & Bit Acc \\ 
    \hline \hline
    Clean  & 21.475  & 0.293  & 0.925 & 0.926 \\ 
    Attack & 70.622 \textcolor{green}{(+49.147)}  & 0.229 \textcolor{red}{(-0.64)}  & 0.000 & 0.522 \\ 
    Recovered & 23.927 \textcolor{green}{(+2.452)}  & 0.292 \textcolor{red}{(-0.01)}  & 0.000 & 0.513 \\ 
    \hline
  \end{tabular}
  }
\label{tab:mir_attack}
\end{table}

\subsection{Mirror Recovered Attack}
\label{sub_sec:Mirror_Recovered_Attack}
Compared with the negative and translation transformations, the mirror transformation makes it more difficult for an attacker to identify the exact transformation that causes the image quality degradation when attempting to remove the watermark by replacing the VAE. To demonstrate the robustness of our method, we assume that the attacker has full knowledge of the mirror transformation process and performs a recovery operation to restore image quality. As shown in the Table~\ref{tab:mir_attack}, even under this most favorable condition for the attacker, the FID remains 2.452 higher than the original value, indicating that the image quality cannot be fully recovered even with complete knowledge of the mirror transformation parameters.

\subsection{Collusion Attack}
\label{sub_sec:collusion_attack}
Inspired by classic collusion attack methods (\eg, SCA~\cite{xiao2022sca}, Byzantine~\cite{fang2020local}, FoolsGold~\cite{fung2018mitigating}), we further evaluate FedOT against a collusion attack tailored to our setting, where multiple malicious clients average their watermarked VAE parameters in an attempt to remove the embedded watermark.

As shown in Table~\ref{tab:collusion_fedotneg_supp}, collusion among 2 or 3 clients degrades the tracing chunk bit accuracy substantially (by 0.216 and 0.321, respectively), while the ownership chunk bit accuracy remains largely unaffected, even slightly improving. This indicates that collusion attacks degrade the tracing watermark but fail to eliminate the ownership watermark. We attribute this to the chunked watermark design: since the ownership bits are shared identically across all clients, averaging colluding clients' parameters reinforces rather than cancels this shared signal, whereas the client-unique tracing bits are diluted by averaging. This distinct degradation pattern provides a reliable signal for identifying colluding participants, which we leave for further discussion in future work.

\begin{table*}[t]
\centering
\caption{Bit accuracy under collusion attacks for FedOT$_\text{neg}$, evaluated against each individual client's watermark.}
\vspace{-1em}
\small
\resizebox{\textwidth}{!}{
\begin{tabular}{c||c|cc|cccc}
\hline
\rowcolor[HTML]{EFEFEF} 
\textbf{Setting} & \textbf{Compared to} & \textbf{FID}$\downarrow$ & \textbf{CLIP-Score}$\uparrow$ & \textbf{Bit Acc}$\uparrow$ & \textbf{Ownership Bit Acc}$\uparrow$ & \textbf{Tracing Bit Acc}$\uparrow$ \\ \hline \hline
\multirow{3}{*}{No Attack} 
    & Client1 & 20.579 & 0.295 & 0.915 & 0.936 & 0.905 \\
    & Client2 & 20.823 & 0.295 & 0.901 & 0.945 & 0.881 \\
    & Client3 & 20.801 & 0.295 & 0.911 & 0.942 & 0.895 \\
\hline
\multirow{2}{*}{2-Client Collusion} 
    & Client1 & \multirow{2}{*}{20.838} & \multirow{2}{*}{0.295} & 0.777 & 0.954 & 0.689 \\
    & Client2 & & & 0.772 & 0.954 & 0.681 \\
\hline
\multirow{3}{*}{3-Client Collusion} 
    & Client1 & \multirow{3}{*}{20.973} & \multirow{3}{*}{0.294} & 0.710 & 0.961 & 0.584 \\
    & Client2 & & & 0.774 & 0.961 & 0.680 \\
    & Client3 & & & 0.756 & 0.961 & 0.653 \\
\hline
\end{tabular}
}
\vspace{-1.25em}
\label{tab:collusion_fedotneg_supp}
\end{table*}

\section{More Experimental Results}
\label{sec:more_experimental_results}

\begin{table*}[t]
\centering
\caption{On the Flickr30K dataset, using 256×256 images and 48-bit watermarks, we compare the generation quality against Stable Signature*. The table on the left reports the results before the VAE replacement attack, while the table on the right shows the performance after the attack.}
\vspace{-1em}
\small
\resizebox{\textwidth}{!}{
\begin{tabular}{c||cccc|cccc}
\hline
\rowcolor[HTML]{EFEFEF} 
\cellcolor[HTML]{EFEFEF} & \multicolumn{4}{c|}{\textbf{Original}} & \multicolumn{4}{c}{\textbf{VAE Replacement Attack}} \\ 
\cline{2-9}
\rowcolor[HTML]{EFEFEF} 
\multirow{-2}{*}{\textbf{Method}} & FID$\downarrow$ & CLIP-Score$\uparrow$ & Detection$\uparrow$ & Bit Acc$\uparrow$ 
                         & FID$\uparrow$ & CLIP-Score$\downarrow$ & Detection$\uparrow$ & Bit Acc$\uparrow$ \\ \hline \hline

\multicolumn{1}{c||}{Original SD} & 22.064 & 0.319 & -- & -- & -- & -- & -- & -- \\
 Stable Signature* & 22.202 & 0.327 & 0.999 & 0.991 &  22.044 \textcolor{red}{(-0.158)} & 0.326 \textcolor{red}{(-0.001)} & 0.000 & 0.500 \\
 FedOT$_\text{w/o LVT}$ & 23.371 & 0.330 & 0.999 & 0.996 & 22.584 \textcolor{red}{(-0.787)} & 0.326 \textcolor{red}{(-0.004)} & 0.000 & 0.499 \\
 $\text{FedOT}_\text{rand}$ & 50.916 & 0.301 & 0.999 & 0.996 & 116.298 \textcolor{green}{(+65.382)} & 0.285 \textcolor{red}{(-0.016)} & 0.000 & 0.491 \\
 \hline
 $\text{FedOT}_\text{tran}$ & 27.814 & 0.317 & 0.999 & 0.986 & 24.427 \textcolor{red}{(-3.387)} & 0.317  & 0.000 & 0.514 \\
 $\text{FedOT}_\text{mir}$ & 28.265 & 0.301 & 0.996 & 0.982 & 125.971 \textcolor{green}{(+97.706)} & 0.232 \textcolor{red}{(-0.069)} & 0.000 & 0.523 \\
 $\text{FedOT}_\text{neg}$ & 27.718 & 0.304 & 0.998 & 0.987 & 71.269 \textcolor{green}{(+43.551)} & 0.267 \textcolor{red}{(-0.037)} & 0.000 & 0.535 \\ \hline
\end{tabular}
}
\vspace{-1.25em}
\label{tab:main_sup_table}
\end{table*}

\subsection{Results on Flicker30K Dataset}
\label{sub_sec:Flicker30K_Dataset}
We conduct additional experiments on Flicker30K~\cite{flicker} dataset using exactly the same experimental configuration as in Section {4.1} of the main paper, except that the translation coefficient is set to 5. The results are shown in Table~\ref{tab:main_sup_table}. As expected, the results of Stable Signature* and $\text{FedOT}_{\text{w/o LVT}}$ are consistent with Table~{1} in the main paper: after the VAE replacement attack, the quality of the generated images remains largely unchanged, but the embedded watermark is completely removed.

When applying the LVT-based methods, VAE replacement leads to noticeable degradation in image quality for $\text{FedOT}_\text{rand}$, $\text{FedOT}_\text{mir}$, and $\text{FedOT}_\text{neg}$. Interestingly, on the Flicker30K dataset, $\text{FedOT}_\text{tran}$ shows a slight improvement in image quality after the attack. This suggests that the translation operation introduces relatively mild perturbations to the original latent space, keeping the transformed latent distribution close to the original one—thereby occasionally producing marginally better images.

Moreover, Flicker30K is approximately three times larger than the Laion10K dataset~\cite{laion_10k} used in the main paper. Although watermark embedding is performed before federated fine-tuning and the VAE is frozen during U-Net updates, the larger dataset improves the U-Net’s generative capability during fine-tuning, resulting in more stable and higher-quality outputs. Since watermark extraction relies on signals embedded in the VAE latent space, improved generation quality indirectly contributes to higher watermark extraction accuracy.

\subsection{VAE Image Reconstruction Quality}
\label{sub_sec:VAE_image_reconstruction_quality}
Our goal is to fine-tune the VAE with LVT so that its latent space undergoes specific transformations within the intersection of the original and transformed latent spaces, thereby enhancing the dependency of the client-trained U-Net on the new VAE. We evaluate the reconstruction quality of the trained VAE using FID, PSNR, and SSIM. As shown in Table~\ref{tab:vae_comparison}, while latent space transformations slightly increase FID, they also improve PSNR and SSIM. A comprehensive analysis suggests that the performance of $\text{FedOT}_\text{rand}$ is the worst. Since the transformation in LVT learns random noise, it has a significant impact on reconstruction, with an FID as high as 14.865. In contrast, $\text{FedOT}_\text{neg}$ achieves the best PSNR and SSIM, and its FID is the lowest compared to the other two FedOT methods. Overall, $\text{FedOT}_\text{neg}$ shows the best reconstruction performance in LVT.

\subsection{Image Generation Quality}
\label{sub_sec:watermark_image_generation_quality}
\begin{figure}[t]
\centering
\includegraphics[width=\textwidth]{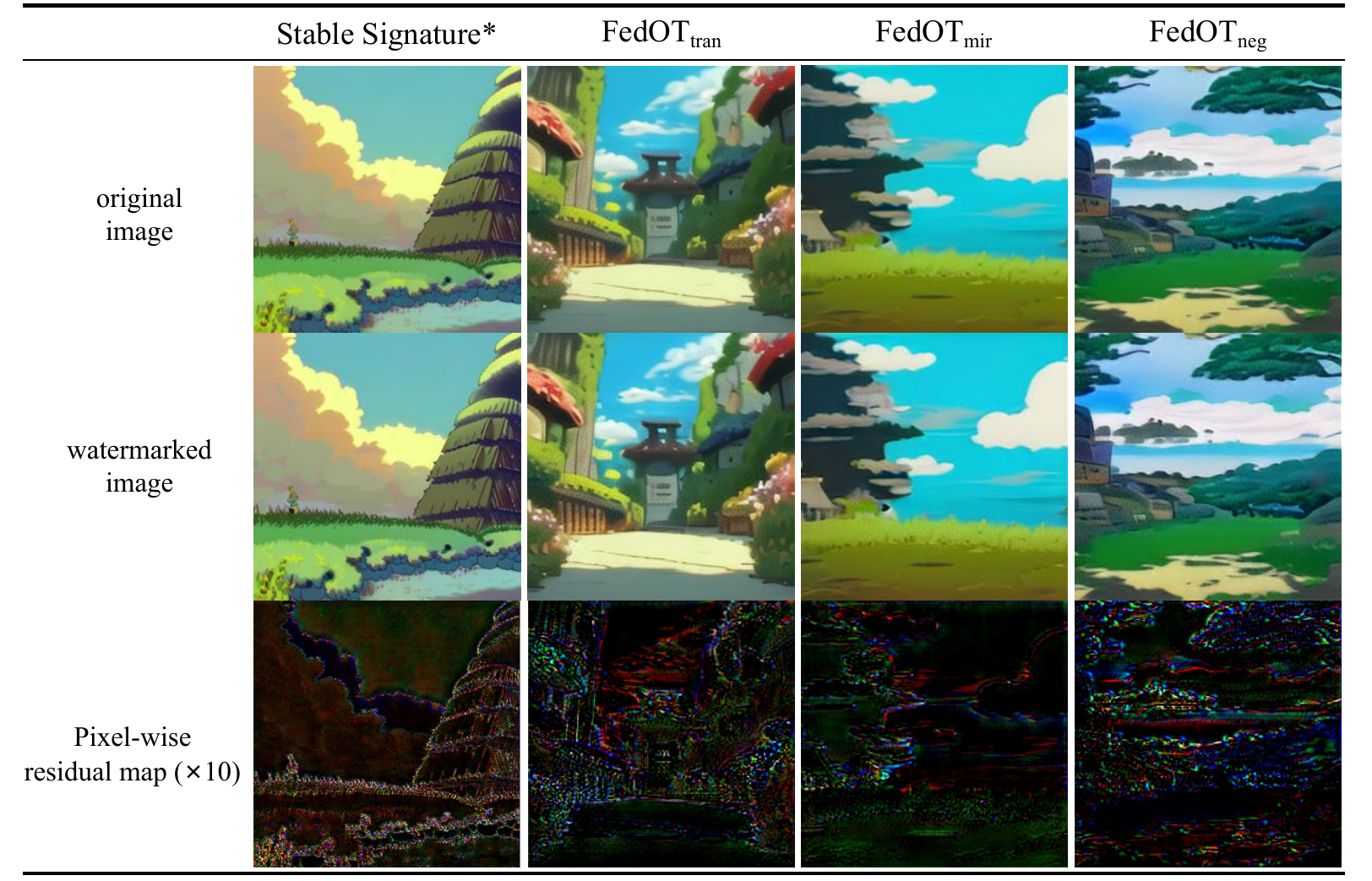}
\vspace{-2em}
    \caption{Pixel-level residual maps between original and watermarked images.}
    \label{fig:residual_map}
\end{figure}

To further evaluate the imperceptibility of watermark embedding, we compare the image quality before and after watermarking across different FedOT variants using PSNR, SSIM, and LPIPS metrics. As shown in Table~\ref{table:psnr}, all FedOT variants achieve high reconstruction fidelity, with PSNR values exceeding 31 dB and SSIM above 0.94, indicating that the watermark introduction causes minimal perceptual distortion. Notably, $\text{FedOT}_\text{tran}$ achieves the best PSNR (33.611) and SSIM (0.968), outperforming even the centralized Stable Signature baseline. $\text{FedOT}_\text{neg}$ shows slightly lower scores, which is consistent with its more aggressive watermark injection strategy. As shown in Figure~\ref{fig:residual_map}, the residual maps further confirm that the pixel-level differences between watermarked and original images are visually imperceptible, demonstrating that FedOT preserves image quality while successfully embedding traceable watermarks.

\begin{table}[t]
\centering
\caption{Presents a comparison of the reconstruction performance between the original VAE and the fine-tuned VAE using four different FedOT methods.}
\vspace{-1em}
\resizebox{0.45\columnwidth}{!}{
  \begin{tabular}{c||ccc}
    \hline
    \rowcolor[HTML]{EFEFEF} 
    \textbf{Method} & FID $\downarrow$ & PSNR $\uparrow$ & SSIM $\uparrow$ \\ 
    \hline \hline
    Original SD  & 1.982  & 26.537  & 0.837 \\ 
    $\text{FedOT}_\text{rand}$ & 14.865 & 26.304  & 0.813 \\ 
    $\text{FedOT}_\text{tran}$ & 4.630  & \textbf{28.107}  & \textbf{0.861} \\ 
    $\text{FedOT}_\text{mir}$ & \textbf{3.867}  & 27.668  & 0.852 \\ 
    $\text{FedOT}_\text{neg}$ & 3.877  & 27.855  & 0.857 \\ 
    \hline
  \end{tabular}
  }
\label{tab:vae_comparison}
\end{table}

\begin{table}[t]
\centering
\caption{PSNR, SSIM, and LPIPS before and after watermark embedding.}
\vspace{-1em}
\resizebox{0.45\columnwidth}{!}{
  \begin{tabular}{c||ccc}
    \hline
    \rowcolor[HTML]{EFEFEF} 
    \textbf{Method} & PSNR $\uparrow$ & SSIM $\uparrow$ & LPIPS $\downarrow$ \\ \hline \hline
    Stable Signature* & 32.801 & 0.963 & 0.009 \\
FedOT$_\text{tran}$ & 33.611 & 0.968 & 0.014 \\
FedOT$_\text{mir}$ & 32.790 & 0.966 & 0.013 \\
FedOT$_\text{neg}$ & 31.042 & 0.949 & 0.018 \\
    \hline
  \end{tabular}
  }
\vspace{-1em}
\label{table:psnr}
\end{table}

\subsection{Selection of Detection Threshold $\tau$}
\label{sub_sec:tau_detection}
\begin{figure}[t]
\centering
\includegraphics[width=0.6\columnwidth]{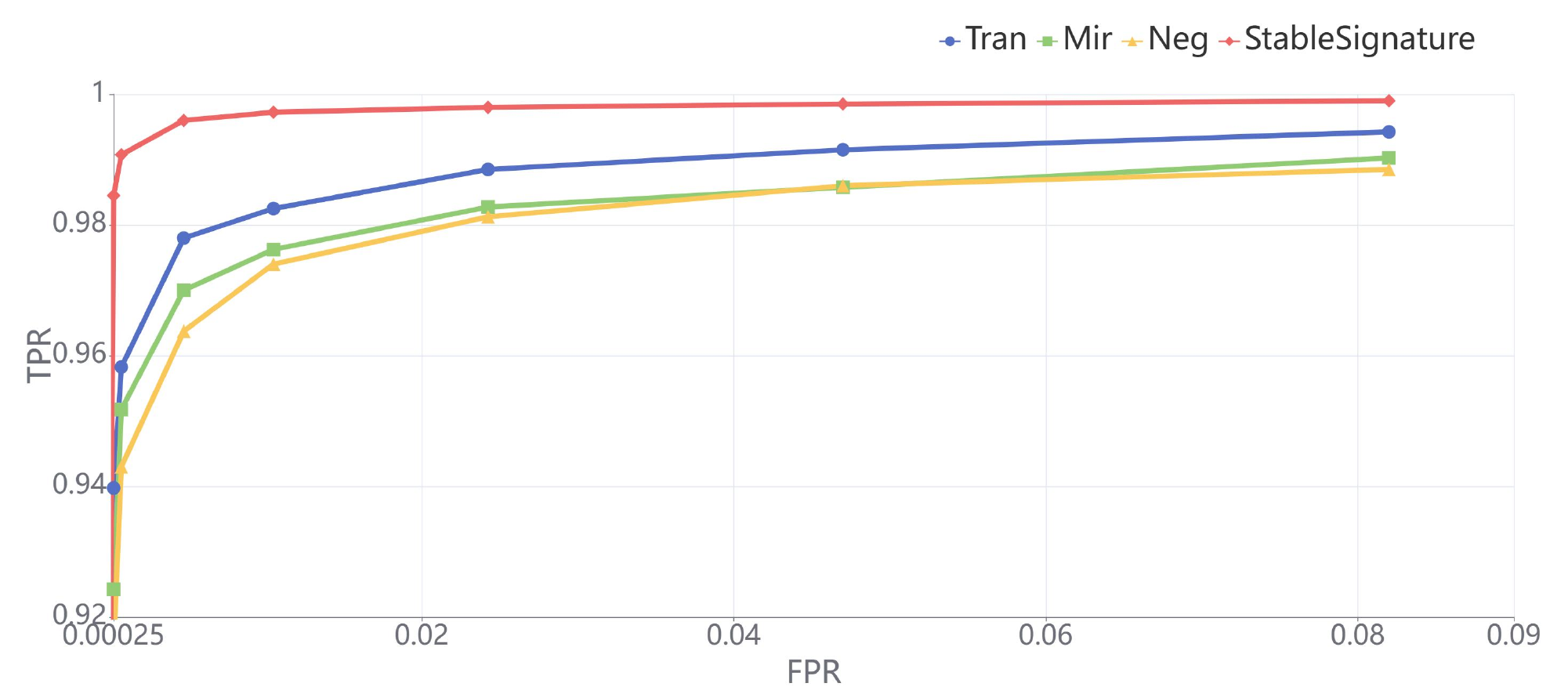}
\vspace{-1em}
    \caption{ROC curves under different detection thresholds $\tau$ across FedOT variants.}
    \label{fig:roc}
\end{figure}

To determine the optimal detection threshold $\tau$, we evaluate the receiver operating characteristic (ROC) curve by recording the True Positive Rate (TPR) and False Positive Rate (FPR) under varying values of $\tau$. As shown in Figure~\ref{fig:roc}, the ROC curve demonstrates strong discrimination ability across all FedOT variants. We select $\tau = 0.69$ as the operating threshold, at which the FPR remains as low as 0.1\%, ensuring that non-watermarked images are rarely misclassified as watermarked while maintaining a high detection rate.

\subsection{End-to-End Attribution Accuracy.}
\label{sec:end-to-end}
To further evaluate the practical tracing capability of FedOT, we sample 1,000 generated images per client and perform end-to-end attribution: for each image, we extract the embedded watermark and identify its source client. As shown in Table~\ref{tab:attribution_supp}, FedOT achieves a per-client attribution accuracy ranging from 96.80\% to 98.70\% across 5 clients (C1--C5), with an overall attribution accuracy of 98.12\% and a false accusation rate of only 1.88\%, demonstrating reliable client-level traceability in practice.

\begin{table*}[t]
\centering
\caption{End-to-end attribution accuracy across 5 clients, sampling 1,000 generated images per client.}
\vspace{-1em}
\small
\begin{tabular}{ccccc||cc}
\hline
\rowcolor[HTML]{EFEFEF} 
\textbf{C1} & \textbf{C2} & \textbf{C3} & \textbf{C4} & \textbf{C5} & \textbf{Acc}$\uparrow$ & \textbf{FAR}$\downarrow$ \\ \hline \hline
96.80\% & 98.10\% & 98.70\% & 98.70\% & 98.30\% & 98.12\% & 1.88\% \\ \hline
\end{tabular}
\label{tab:attribution_supp}
\end{table*}

\subsection{Additional Visualizations}
\label{sub_sec:More_FedOT_visual_results}
FedOT demonstrates strong robustness against VAE replacement attacks. When an adversary attempts to remove the watermark by replacing the VAE, FedOT ensures that high-quality image generation becomes infeasible. Malicious attackers may exploit the compromised model to generate various images for profit. In the following visualizations, the translation coefficient is set to 5.

\begin{figure*}[t]
\centering
\includegraphics[width=\textwidth]{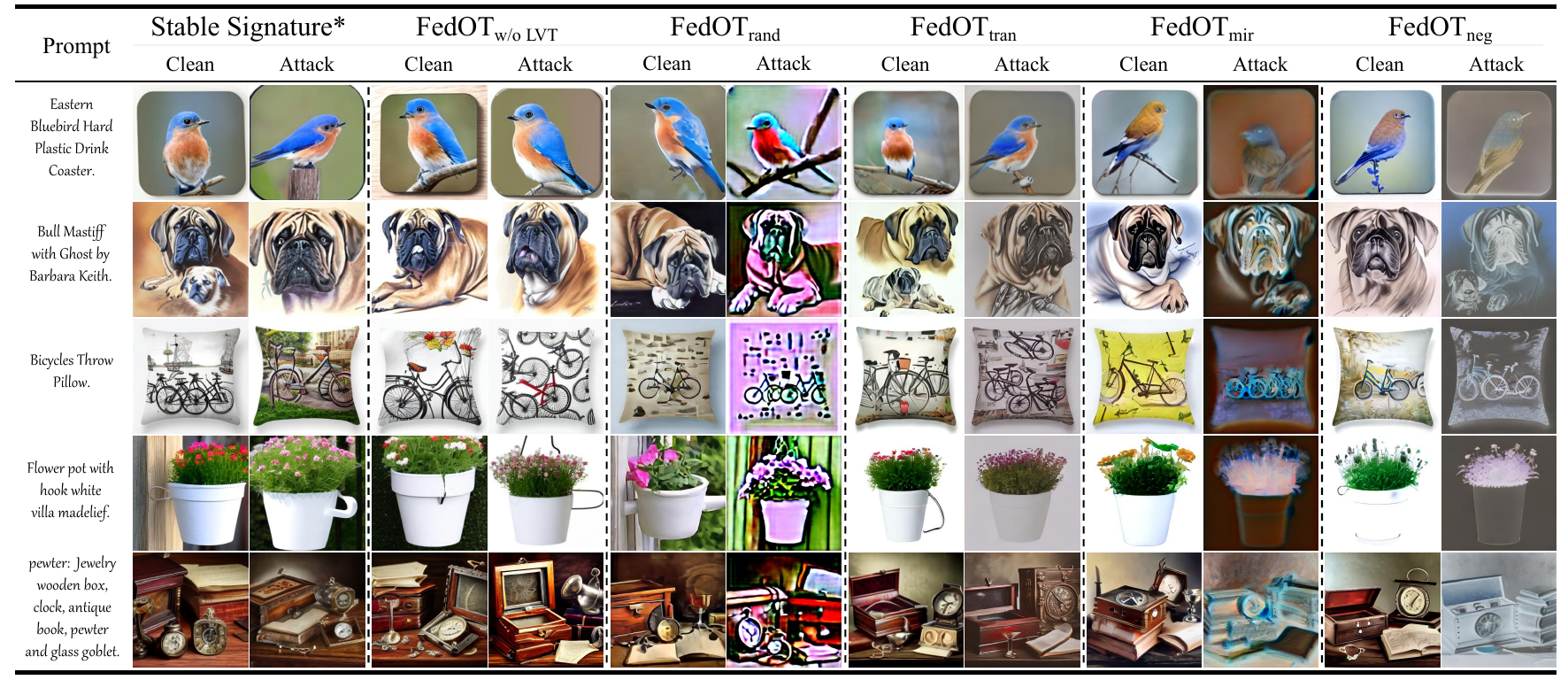}
\vspace{-2em}
    \caption{Additional generated image results under FedOT include various animals, landscapes, and paintings, using prompts from the Laion10K dataset.}
    \label{fig:results1}
\vspace{-1.5em}
\end{figure*}

\begin{figure*}[t]
\centering

\includegraphics[width=\textwidth]{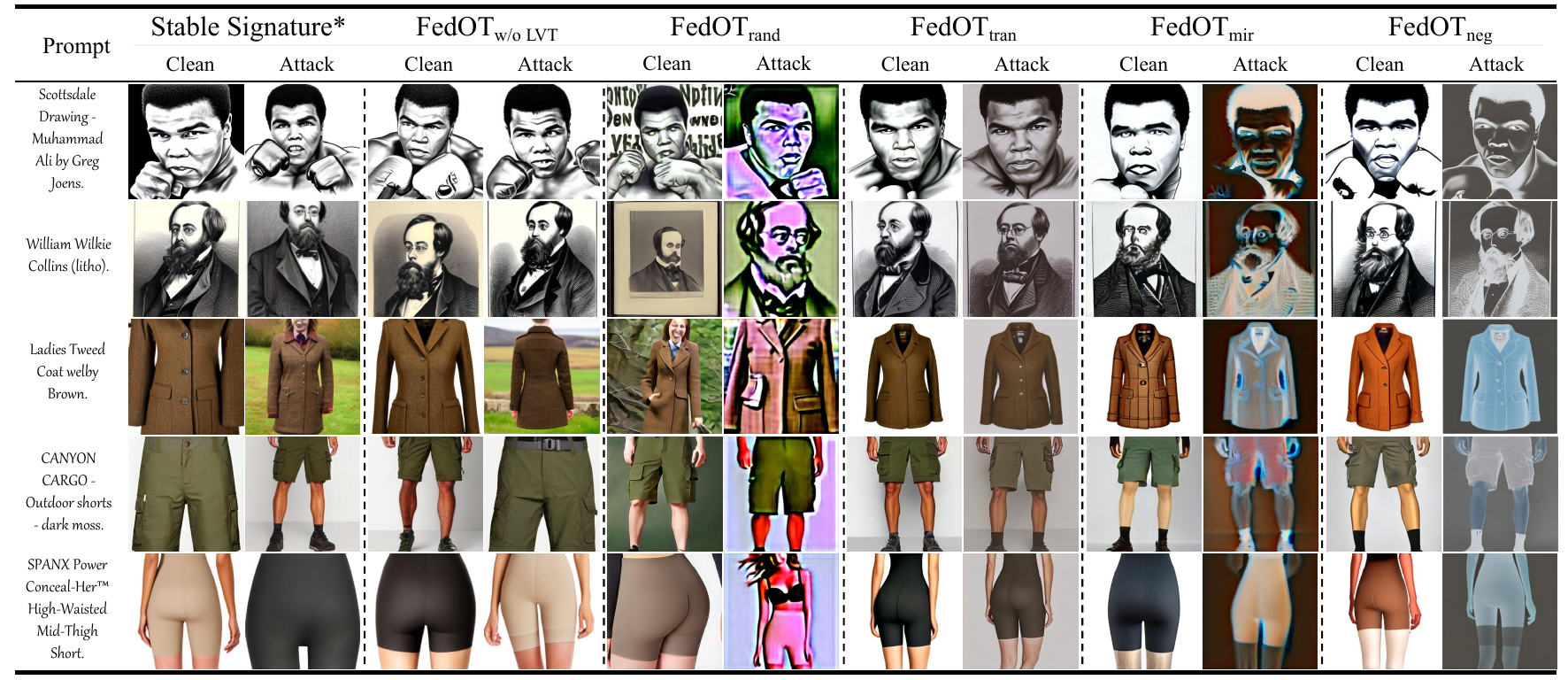}
\vspace{-2em}
    \caption{Additional generated image results under FedOT include various human-related images, using prompts from the Laion10K dataset. }
\vspace{-1em}
\label{fig:results2}
\end{figure*}

To illustrate this, Fig.~\ref{fig:results1} presents examples of animal, landscape, and oil painting images, while Fig.~\ref{fig:results2} showcases images related to humans. The leakage of such models often raises serious privacy concerns, allowing attackers to profit from unauthorized use. In particular, human-related images frequently involve sensitive information, making model leakage detection and tracing even more crucial.

This highlights the significance of FedOT. By embedding robust watermarking into federated diffusion models, FedOT not only deters unauthorized use but also ensures that attempts to bypass tracking lead to significant image quality degradation. Given the privacy risks associated with model leakage, especially in the context of human-related images, our approach offers an effective solution for ownership verification and model tracing, enhancing accountability and security in federated generative learning.

\vspace{-0.5em}

\section{Additional Discussions}
\subsection{Secrecy of LVT}
\label{sub_sec:Secrecy_of_the_LVT}
In our design, the transformation $T$ is kept confidential. Only the server is aware of the specific transformation applied to the VAE, while clients remain unaware that the latent space of the VAE has been modified. Even if a malicious client attempts to remove the watermark by replacing the VAE and discovers that the latent space has undergone some transformation, it is still difficult to precisely infer $T$. This is because, during the training of the LVT, the learned VAE parameters approximate the designed mapping but inevitably include minor deviations. These deviations further increase the difficulty for an attacker to accurately reverse-engineer the transformation.

As demonstrated in the experiments presented in Appendix~\ref{sub_sec:Negative_Recovered_Attack}--\ref{sub_sec:Mirror_Recovered_Attack}, even under the most favorable conditions for the attacker, attempting to recover the latent space of the VAE still leads to low-quality image reconstruction. These results further confirm the robustness of our proposed method.

\subsection{Limitations}
\label{sub_sec:limitation}
One limitation of watermark embedding techniques, including our approach, is that the introduction of watermarks can lead to a slight decrease in image generation quality. This is a common trade-off in watermarking methods, where preserving the integrity of the watermark often comes at the expense of some minor degradation in image quality. While our method focuses on minimizing this impact, the trade-off remains an inherent challenge when embedding robust watermarks into generative models. Future work may explore ways to optimize watermarking techniques to further reduce the impact on generation quality.

\subsection{Communication Overhead}
\label{sub_sec:Communication_overhead}
In the entire federated training process, the complete LDM model is transmitted only once during the initial distribution. For all subsequent communication rounds, only the U-Net parameters are exchanged between the server and clients. As a result, the communication overhead introduced by other components is incurred only once and does not contribute to repeated transmission costs. This design significantly reduces the communication burden and improves training efficiency.

\subsection{Scalability}
\label{sub_sec:Scalability}
To support a larger number of clients, the server maintains one uniquely watermarked VAE for each client, with each model occupying approximately 335MB of storage. This leads to a linearly increasing storage requirement as the number of clients grows. Due to hardware limitations, we are not able to scale to millions of clients. However, since our watermarking method is similar to that of Stable Signature\cite{Stable_signature}, their findings provide indirect support for our approach, as they have shown that watermarking remains effective even with up to $10^7$ users.

\end{document}